\documentclass[lettersize,journal]{IEEEtran}
\usepackage{amsmath,amsfonts}
\usepackage{algorithmic}
\usepackage{array}
\usepackage{subfloat}

\usepackage{textcomp}
\usepackage{stfloats}
\usepackage{url}
\usepackage{verbatim}
\usepackage{graphicx}
\def\BibTeX{{\rm B\kern-.05em{\sc i\kern-.025em b}\kern-.08em
    T\kern-.1667em\lower.7ex\hbox{E}\kern-.125emX}}
\usepackage{balance}

\usepackage[colorlinks,urlcolor=blue,linkcolor=blue,citecolor=blue]{hyperref}
\usepackage{color}
\definecolor{Gray}{rgb}{0.572,0.572,0.572}
\usepackage{color,array}
\usepackage{tablefootnote}
\usepackage{threeparttable}

\usepackage{algorithm}
\usepackage{subcaption}

\usepackage{cite}
\usepackage{romannum}
\usepackage{diagbox}
\usepackage{makecell}
\usepackage{multirow}
\usepackage{ragged2e} 
\usepackage{float}
\usepackage{enumitem}
\usepackage{tabularray}
\usepackage{lscape}
\usepackage{multirow}
\usepackage{longtable}
\begin{document}

\title{Proximity-Based Evidence Retrieval for\\ Uncertainty-Aware Neural Networks}

\author{Hassan Gharoun\textsuperscript{1}, Mohammad Sadegh Khorshidi\textsuperscript{1}, Kasra Ranjbarigderi\textsuperscript{1}, Fang Chen\textsuperscript{1}, and Amir H. Gandomi\textsuperscript{1,2,3}
\thanks{\textsuperscript{1}Faculty of Engineering \& IT, University of Technology Sydney, e-mails: Hassan.Gharoun@student.uts.edu.au,  Mohammadsadegh.khorshidialikordi@student.uts.edu.au, Kasra.Ranjbarigderi@student.uts.edu.au, Fang.Chen@uts.edu.au, Gandomi@uts.edu.au.}
\thanks{\textsuperscript{2}University Research and Innovation Center (EKIK), Óbuda University.}
\thanks{\textsuperscript{3}Corresponding author}}

\maketitle

\begin{abstract}
This work proposes an evidence-retrieval mechanism for uncertainty-aware decision-making that replaces a single global cutoff with an evidence-conditioned, instance-adaptive criterion. For each test instance, proximal exemplars are retrieved in an embedding space; their predictive distributions are fused via Dempster–Shafer theory. The resulting fused belief acts as a per-instance thresholding mechanism. Because the supporting evidences are explicit, decisions are transparent and auditable. Experiments on CIFAR-10/100 with BiT and ViT backbones show higher or comparable uncertainty-aware performance with materially fewer confidently incorrect outcomes and a sustainable review load compared with applying threshold on prediction entropy. Notably, only a few evidences are sufficient to realize these gains; increasing the evidence set yields only modest changes. These results indicate that evidence-conditioned tagging provides a more reliable and interpretable alternative to fixed prediction entropy thresholds for operational uncertainty-aware decision-making.
\end{abstract}


\section{Introduction}
\IEEEPARstart{I}{n} the landscape of modern artificial intelligence (AI), the pursuit of predictive accuracy has driven neural networks (NNs) to achieve superhuman performance across a multitude of domains. However, in many real-world applications, particularly those with high stakes, a correct prediction is only part of the requirement.

This is crucial because most conventional machine learning (ML) models issue single-point predictions. In particular, NNs typically output class probabilities through a softmax layer, which represent only a deterministic point estimate conditioned on the model’s fixed parameters and training data. These probabilities reflect the model’s relative preference among classes given its fixed state after training. High probability does not necessarily imply that the prediction is reliable. This is where uncertainty quantification (UQ) methods emerges as a critical paradigm. UQ techniques, ranging from theoretically principled Bayesian methods to more pragmatic techniques such as ensemble \cite{abdar2021review}, generate a distribution of outputs for each instance rather than a single deterministic prediction. Thus, they go beyond the raw probability assigned to a class, enable to quantify how much the model can be trusted in light of variability in predictions in presence of various uncertainty sources. This distinction is critical: while what the model predicts is described by probabilities, how much trust should be placed in those predictions is informed by uncertainty measures.

Measuring uncertainty—i.e., the variability of output distributions obtained through UQ methods—and embedding it into the decision-making loop can significantly enhance the reliability and safety of AI systems. Such integration allows models to recognize when predictions should be deferred, thereby involving human collaborators or triggering safety protocols in high-stakes environments.

Prediction entropy (PE) is one of the most widely used measures of predictive uncertainty. It is computed directly from the class probability distribution formulated by Eq.~\ref{Eq:PE}:

\begin{equation}
    PE(\mathbf{x}) = - \sum_{c=1}^{C} \mu_{\text{pred}} (\mathbf{x}, c) \log[\mu_{\text{pred}} (\mathbf{x}, c)]
    \label{Eq:PE}
\end{equation}

\noindent where $\mu_{\text{pred}}(\mathbf{x}, c)$ is the averaged prediction for class $c$ over input $\mathbf{x}$.

\noindent A low PE indicates that the predictive distribution is sharply concentrated on a single class, meaning that the stochastic predictions obtained from the UQ method are consistent across multiple runs. In this case, the model exhibits low variability and its output can be regarded as certain and stable. Conversely, a high PE reflects a more dispersed predictive distribution, where probability mass is spread across several classes. This implies that the stochastic predictions fluctuate significantly across runs, signaling high variability and, consequently, greater uncertainty in the model's decision. A common approach in the literature applies a threshold-based rule on PE, where instances with entropy values above a chosen cutoff are flagged as uncertain, while those below the threshold are treated as certain. However, this strategy is not without limitations.

Consider the binary classification case, where:

\begin{equation}
\begin{split}
    PE(\mathbf{x}) &= - \mu_{\text{pred}}(\mathbf{x},C_1)\log\!\big(\mu_{\text{pred}}(\mathbf{x},C_1)\big) \\
                    & ~~~- \mu_{\text{pred}}(\mathbf{x},C_2)\log\!\big(\mu_{\text{pred}}(\mathbf{x},C_2)\big)
\end{split}
\end{equation}

\noindent with $\mu_{\text{pred}}(\mathbf{x},C_1)+\mu_{\text{pred}}(\mathbf{x},C_2)\approx 1$. 
By direct inspection of the entropy function, the maximum is attained when $\mu_{\text{pred}}(\mathbf{x},C_1)=0.5$, giving $PE_{\max}=1$ under log base 2.

For a prediction to lie below a small entropy threshold $\tau$ (with $\tau \ll PE_{\max}$), the probability assigned to the most likely class must be close to one. 
For instance, in the binary case with $\tau=0.1$ and log base 2, solving $PE<=0.1$ numerically yields $\mu_{\text{pred}}(\mathbf{x},C_1)\approx 0.933$. 
This illustrates that low entropy values can only be achieved when the average prediction is highly concentrated on a single class.  

The same principle extends to multiclass settings: as the number of classes $C$ increases, the maximum entropy becomes $\log(C)$, and meeting a fixed low threshold $\tau$ requires even greater concentration of probability mass in the predicted class. 
Hence, the mathematical form of entropy inherently favors sharp probability distributions, meaning that predictions are only labeled as ``low uncertainty'' when the mean predicted probabilities are extremely peaked—regardless of whether such concentrated confidence is actually warranted by the underlying data. As a consequence, many correct predictions may still be classified as uncertain whenever their probability distributions are less sharply concentrated, even though the predicted class matches the ground truth.

To address the limitations of threshold-based entropy methods, this study proposes a more adaptive approach toward evidence-based uncertainty-aware decision-making. Instead of the question \textit{"Is this prediction uncertain according to a fixed threshold?"} being asked, the question \textit{"Given similar historical cases, how should this uncertainty be interpreted?"} is posed. This is analogous to the processes of human intelligence in categorization, as outlined in Nosofsky’s exemplar theory \cite{nosofsky2011generalized}. According to exemplar theory \cite{nosofsky2011generalized}, when humans encounter stimuli that must be categorized—such as determining whether a sound belongs to a familiar category or whether a visual pattern matches a known type—cognitive processing operates through similarity-based comparisons with stored examples. In these models, the mind computes the similarity between the current stimulus and previously encountered instances across multiple relevant dimensions. A categorization decision then emerges from the accumulated similarity values, weighted by the strength with which each exemplar is stored in memory. The relative strength of evidence favoring each potential category determines the likelihood of a given response.

Accordingly, this study proposes a method in which uncertainty assessment is performed by combining proximity-based evidences stored in a historical reference set, referred to as the evidence set. For each new instance, similar examples are retrieved from the evidence set, where similarity is defined in terms of proximity in the embedding space. These retrieved examples act as evidential references, and their predictive distributions provide multiple perspectives on variability. An uncertainty-aware decision is then made by evaluating whether the new instance exhibits the same uncertainty behavior as its similar evidential counterparts. In this way, the decision process becomes more adaptive: rather than applying a fixed rule uniformly to all instances, the model determines whether the model demonstrates consistent behavior across both the current prediction and its closest historical examples.

The remainder of this paper is organized as follows. Section \ref{sec:Background} summarizes prior work on uncertainty quantification, and its role in decision-making frameworks. Section \ref{sec:Methodology} introduces the proposed approach in detail. Section \ref{sec:Experiments} describes the experimental setup, covering datasets, model architectures, evaluation protocols, and implementation specifics. Section \ref{sec:Results} reports and discusses the findings, emphasizing the effects of the proposed method on decision reliability. Finally, Section \ref{sec:Conclusion} closes the paper with concluding remarks and suggests directions for future investigations.

\section{Background} \label{sec:Background}
UQ is the science of determining how likely certain outcomes are when some aspects of a system are not exactly known. A central tenet of modern UQ is the recognition that not all uncertainty is of the same nature. A critical distinction is drawn between two fundamental types of uncertainty \cite{abdar2021review}: aleatoric and epistemic. Aleatoric uncertainty refers to the inherent randomness or stochasticity within a system or a measurement process. It is the variability that would remain even with perfect knowledge of the system's parameters and governing equations. This type of uncertainty is often described as irreducible noise. Epistemic uncertainty arises from a lack of knowledge about the system. This form of uncertainty is, in principle, reducible with the acquisition of more data or the development of more accurate models.

The Bayesian framework, therefore, provides a direct and intuitive mechanism for quantifying epistemic uncertainty. The Bayesian framework defines probability as a degree of belief or a measure of certainty about a proposition, given a state of knowledge. This is an epistemic view of probability. From this perspective, it is perfectly coherent to assign a probability distribution to a model parameters -often called posterior distributions—that describe the degree of (un)certainty associated with each unknown, explicitly incorporating both prior knowledge and observed evidence \cite{neal2012bayesian}. Therefore, 
the Bayesian framework allows for probabilistic statements about hypotheses and parameters, such as "there is a 95\% probability that the true parameter value lies within this interval. 

The literature presents various established Bayesian UQ methods for posterior parameter estimation, including Markov chain Monte Carlo (MCMC) \cite{gamerman2006markov}, Variational Inference (VI) \cite{graves2011practical}, Variational Autoencoders (VAE) \cite{kingma2013auto}, Laplace approximation \cite{ritter2018scalable}, Bayes By Backprop (BBB)\cite{fortunato2017bayesian}, and Bayesian active learning (BAL) \cite{gal2017deep}. Sidestepping computational complexities, Monte Carlo dropout, introduced by Gal and Ghahramani \cite{gal2016dropout}, offers a computationally efficient approximation by leveraging dropout as a variational inference technique. This method provides a practical trade-off between UQ and computational feasibility. Ensemble methods are another UQ approach that focuses on computational efficiency. Ensemble methods train multiple models with different initializations or data subsets and aggregate their predictions to estimate uncertainty \cite{lakshminarayanan2017simple}. Their primary advantage lies in computational tractability—individual models can be trained in parallel, and uncertainty estimation requires only forward passes through the ensemble members.

The literature on uncertainty-aware decision making is replete with studies that adopt these standard UQ methods or present slight modifications thereof to estimate model uncertainty, for instances \cite{aseeri2021uncertainty,martin2024uncertainty,senousy2021mcua,carneiro2020deep,habibpour2021uncertainty,habibpour2023uncertainty,westermann2021using,yao2024uncertainty,shamsi2023novel}. The comprehensive review demonstrates that the most common framework for uncertainty-aware decision-making is the application of a threshold on prediction variability, typically expressed either as PE or as the variance of predictive distributions and applied uniformly across all samples. 

In contrast to these uniform threshold approaches, this study proposes a proximity-based evidence retrieval framework where uncertainty-aware decisions are adaptive and contextually informed. Rather than applying fixed global thresholds to all predictions, the proposed method retrieves semantically similar instances from a curated evidence set and employs Dempster-Shafer theory to fuse their uncertainty representations.

Dempster-Shafer (DS) theory, also known as the theory of belief functions or evidence theory, provides a mathematical framework for reasoning under uncertainty that generalizes classical probability theory \cite{shafer1976mathematical,dempster1967upper}. Unlike traditional probability theory, which requires precise probability assignments, DS theory allows for the explicit representation of ignorance and partial knowledge through the use of belief functions and mass assignments.

The fundamental concept in DS theory is the \emph{frame of discernment} $\Omega$, which represents the set of all possible mutually exclusive and exhaustive hypotheses under consideration. For a binary classification problem, the frame of discernment consists of
$\Omega = \{c_1, c_2\}$. The power set $2^\Omega$ contains all possible subsets of $\Omega$, including the empty set $\emptyset$, the singletons $\{c_1\}, \{c_2\}$, and the whole set $\Omega$ itself ( $2^\Omega = \{\emptyset, \{c_1\}, \{c_2\}, \{c_1, c_2\}\}$) . 

\noindent A \emph{mass function}, also called a \emph{basic belief assignment (BBA)}, is defined as a mapping
$m: 2^\Omega \rightarrow [0,1]$, such that:

\begin{equation}
    m(\emptyset) = 0, \qquad \sum_{A \subseteq \Omega} m(A) = 1.
\end{equation}

\noindent The value $m(A)$ quantifies the exact degree of evidence assigned to subset $A \subseteq \Omega$, without being distributed to any of its strict subsets. 

\noindent For a binary classification problem, evidence is assumed to be allocated in the following way:

\begin{equation}
    m(\{c_1\}) = p_{c_1}, \quad m(\{c_2\}) = p_{c_2}, \quad m(\Omega) = 1 - p_{c_1} - p_{c_2}.
\end{equation}

\noindent In this setting:
\begin{itemize}
    \item $A = \{c_1\}$ is regarded as evidence supporting class $c_1$.
    \item $A = \{c_2\}$ is regarded as evidence supporting class $c_2$.  
    \item $A = \Omega = \{c_1, c_2\}$ is interpreted as the portion of evidence regarded as \emph{undecided} between $c_1$ and $c_2$.  
\end{itemize}

\noindent The subsets $A$ for which $m(A) > 0$ are referred to as the \emph{focal sets}. In this example, the focal sets are given by $\{c_1\}, \{c_2\},$ and $\Omega$.  

\noindent The \emph{plausibility function} $\text{Pl}(A)$ measures the extent to which evidence fails to refute hypothesis $A$:

\begin{equation}
    \text{Pl}(A) = \sum_{B \cap A \neq \emptyset} m(B) = 1 - \text{Bel}(\overline{A})
\end{equation}

\noindent where $\overline{A}$ denotes the complement of $A$. For a binary classification problem under consideration, the plausibility for $A = \{c_1\}$ and $A = \{c_2\}$ is calculated as:

\begin{itemize}
    \item $Pl(\{c_1\}) = m(\{c_1\}) + m(\Omega) = p_{c_1} + (1 - p_{c_1} - p_{c_2}).$
    \item $Pl(\{c_2\}) = m(\{c_2\}) + m(\Omega) = p_{c_2} + (1 - p_{c_1} - p_{c_2}).$
\end{itemize}

\noindent When multiple independent sources of evidence are available, Dempster's rule of combination provides a mechanism for aggregating mass functions. For two mass functions $m_1$ and $m_2$, the combined mass function $m = m_1 \oplus m_2$ is computed as:
\begin{equation}
    m(A) = \frac{\sum_{B \cap C = A} m_1(B) m_2(C)}{1 - K}
\end{equation}
\noindent where $K = \sum_{B \cap C = \emptyset} m_1(B) m_2(C)$ represents the degree of conflict between the two sources of evidences (B, C). The denominator normalizes the result, effectively redistributing the conflicting mass among the non-conflicting elements.

Overally, the integration of DS theory into ML has followed two major trajectories. The first line of research has focused on embedding DS theory within individual classifiers. Early work by \cite{denoeux2000neural} introduced prototype-based evidential neural networks, where input patterns generated BBAs that were fused through Dempster’s rule, with uncertainty represented explicitly as mass assigned to ignorance. This concept was later extended into the deep learning era by \cite{sensoy2018evidential}, who proposed Evidential Deep Learning (EDL). EDL replaces the softmax function with non-negative evidential outputs that parameterize Dirichlet distributions, allowing uncertainty to be decomposed into class-specific belief and residual ignorance in a single deterministic forward pass. Subsequent extensions, such as region-based EDL for medical segmentation \cite{li2023region} and modified EDL with an explicit “unknown” class \cite{nagahama2023learning}, improved robustness, eliminated reliance on arbitrary thresholds, and enabled open-set recognition. Other variants, including evidential CNNs with DS layers \cite{yaghoubi2022cnn, tong2021evidential}, set-valued evidence adjustment strategies \cite{yuan2020evidential}, conflict management strategy with novel correlation belief function \cite{tang2023new}, and conformal threshold strategy on plausibility values \cite{kempkes2024reliable}, have further demonstrated the capacity of DS to predictions that incorporate ignorance, supporting uncertainty-aware decision making.

The second trajectory has applied DS theory at the ensemble and decision-fusion level. In this setting, multiple classifiers or feature views are treated as sources of evidence, and their outputs are transformed into BBAs that are fused using Dempster’s rule or its variants. This strategy has been employed across diverse domains, from biomedical signal processing and image processing \cite{yazdani2009classification,liu2023classifier}, to fault diagnosis \cite{oukhellou2010fault,yaghoubi2022novel}, where heterogeneous classifiers or multimodal sensors naturally provide complementary evidence. Also, in the literature, DS theory has been employed to combine the outputs of classifiers in an ensemble—treated as pieces of evidence— to compute weights for each classifier, thereby replacing standard majority voting with a weighted majority voting scheme \cite{chamlal2025ensemble}. Although the strict independence assumption of Dempster’s rule is rarely satisfied in practice, empirical studies consistently demonstrate that DS fusion offers superior robustness to noise and conflict compared to traditional ensemble strategies.

Across both trajectories, a common advantage of DS-based models is the ability to represent and propagate uncertainty and ignorance explicitly, rather than relying solely on prediction probabilities. Deterministic evidential models provide scalable single-pass uncertainty quantification, while ensemble-based DS fusion captures epistemic variability across models or modalities.

The present work introduces a third trajectory that leverages DS theory for "proximity-based evidence contextualization" in uncertainty-aware decision making. Unlike existing approaches that embed DS theory within individual classifiers or apply it at the ensemble level, the proposed framework treats semantically similar historical instances as independent sources of evidence for each test prediction. This approach addresses fundamental limitations in both trajectories: individual evidential classifiers lack external validation context and may suffer from overconfidence, while ensemble methods aggregate evidence from models trained on the same data distribution without considering instance-specific similarity patterns. By retrieving evidence from a curated set based on learned feature similarity and fusing their uncertainty representations through Dempster's rule, the framework provides adaptive, sample-specific decision criteria that contextualize individual predictions within neighborhoods of semantically similar instances. Here, crucially, DS is not used to determine or change the predicted class: the class label is produced by the classifier model, and DS operates only over the retrieved exemplars’ predictive distributions to quantify beliefs and to issue a certain/uncertain tag.

% This eliminates the need for global threshold tuning that plagues traditional uncertainty-based rejection methods, while maintaining the computational efficiency of single-model inference augmented by fast similarity search. 

\section{Methodology} \label{sec:Methodology}

\begin{figure*}[h]
  \centering
  \captionsetup[subfloat]{font=tiny}
  {\includegraphics[width=0.9\textwidth]{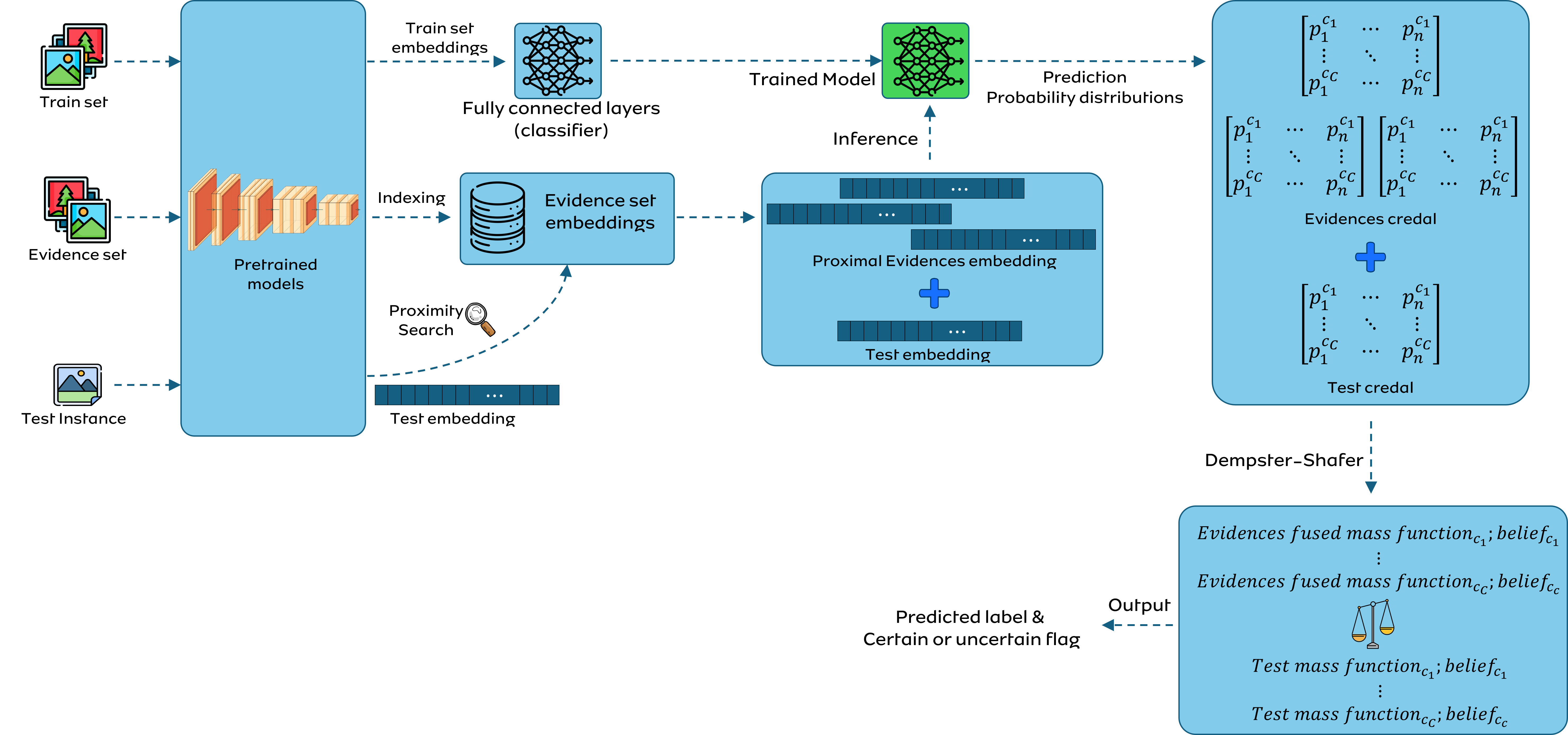}}\par
  \caption{Overview of the proximity-based evidence retrieval method for uncertainty-aware neural network inference. The framework combines individual test sample uncertainty estimates with neighborhood consensus through DS mass function fusion to enable contextual uncertainty-aware decisions.}
  \label{fig:workflow}
\end{figure*}

The proposed framework employs Monte Carlo dropout (MCD) as Bayesian UQ, integrating uncertainty estimates with DS evidence theory through a proximity-based evidence retrieval mechanism to enable uncertainty-aware prediction systems. Figure~\ref{fig:workflow} presents the overall architecture of the proposed method, illustrating the four sequential phases: neural network training with MCD as a base classifier, evidence set construction, proximity-based evidence retrieval, and decision-making through evidence fusion. This approach addresses limitations in traditional entropy-based uncertainty measures by contextualizing prediction confidence within neighborhoods of semantically similar instances. Each component of the framework is detailed in the following subsections.

\subsection{Preliminary: Data Partitioning}
The methodology necessitates a three-way data partitioning approach beyond standard train-test divisions: (1) training set, (2) test set, and (3) evidence set. Following standard procedures, the training set was used exclusively for neural network training, while the test set remained completely unseen during both training and evidence construction phases, ensuring unbiased performance assessment.
The proposed methodology requires an additional evidence set with known ground truth labels to support proximity-based evidence retrieval for uncertainty-aware reasoning without participating in the training process. This evidence set serves as a labeled reference corpus for proximity-based evidence retrieval detailed in subsequent sections \ref{sec:DS}.  

To formalize this strategy mathematically, Let the dataset be split into a \emph{training set} $\mathcal T$, a
\emph{evidence set} $\mathcal E$ and a \emph{test set} $\mathcal X$ with
$\mathcal E\cap(\mathcal T\cup\mathcal X)=\emptyset$.
The neural network classifier $f_{\boldsymbol\theta}\colon\mathbb R^{H\times W\times3}\!\to\!\Delta^{C-1}$ is trained only on $\mathcal T$.

\subsection{Uncertainty Quantification: From Monte Carlo Dropout to Credal Intervals}
A fundamental step in uncertainty-aware decision making is the quantification of model uncertainty. Considering that a neural network is adopted as the core decision-making architecture in this study, MCD is employed as an efficient UQ technique to estimate epistemic uncertainty within a Bayesian approximation framework. While dropout was originally introduced as a regularization method to mitigate overfitting by randomly disabling neurons during training, its application can be extended to the inference stage to enable uncertainty estimation \cite{gal2016dropout}. By keeping dropout active at test time and performing multiple stochastic forward passes, a distribution of outputs is generated for each input. This procedure is theoretically grounded in the Bayesian framework, as each stochastic pass corresponds to sampling from a variational distribution over the weights. As such, it approximates Bayesian inference by capturing epistemic uncertainty through a distribution over model parameters. 
The collection of stochastic predictions obtained through this process is subsequently transformed into credal intervals that provide lower and upper probability bounds for each class. Specifically: 

\noindent For any input $x$, $M$ stochastic forward passes are performed:
\[
  \bigl\{\,\mathbf p_m(x)\in\Delta^{C-1}\bigr\}_{m=1}^{M},
\]
where $\mathbf p_m(x)=f_{\boldsymbol\theta}(x)$ represents the output probability vector from the $m$-th forward pass with dropout enabled, and $\Delta^{C-1}$ denotes the $(C-1)$-dimensional probability simplex.

The collection of Monte Carlo samples $\{p_{m,c}(x)\}_{m=1}^{M}$ for each class $c$ enables the construction of credal intervals via empirical quantile computation:

\begin{equation}
  p^{-}_{c}(x)=\operatorname{quantile}_{0.1}\!\bigl\{p_{m,c}(x)\bigr\}_{m=1}^{M}
\end{equation}
\begin{equation}
  p^{+}_{c}(x)=\operatorname{quantile}_{0.9}\!\bigl\{p_{m,c}(x)\bigr\}_{m=1}^{M}
\end{equation}
These quantiles define the credal interval $[p^{-}_{c}(x),p^{+}_{c}(x)] \subset [0,1]$ that characterizes the uncertainty in the model's predicted probability for class $c$.

\subsection{Mapping Intervals to Dempster–Shafer Mass} \label{sec:DS}
Following credal interval construction, the next step involves transforming these uncertainty representations into DS mass functions to enable evidence-theoretic reasoning. This conversion follows established protocols for mapping credal intervals to belief structures within the DS framework.
The transformation leverages the lower bounds of credal intervals as direct evidence for specific classes, while the remaining probability mass represents epistemic uncertainty. In DS theory, mass functions should reflect the strength of evidence rather than precise probabilities. The lower bound captures the "committed" or "certain" belief that can be attributed to each class, while the interval width (upper - lower) represents epistemic uncertainty that gets allocated to ignorance.

Formally, every test sample $x^*$ is associated with a mass function defined as Eq. \ref{Eq:mass}:

\begin{equation}
    m_{x^*}(c)=p^{-}{c}(x^*),
    \qquad
    m_{x^*}(\Omega)=1-\sum_{c=1}^{C} p^{-}_{c}(x^*),
\label{Eq:mass}
\end{equation}

This construction enables each evidence sample to contribute both specific class beliefs (through singleton masses) and ignorance mass to subsequent evidence fusion processes.

\subsection{Proximity-Based Evidence Retrieval and Dempster-Shafer Evidence Fusion}

Evidence retrieval operates through similarity search in the learned representation space extracted from pretrained neural network architectures trained on ImageNet. The penultimate layer activations serve as feature representations that capture visual patterns learned during large-scale pretraining. Under the assumption that proximity in this learned feature space correlates with visual similarity, cosine similarity is employed to identify instances with potentially related visual characteristics for evidence aggregation.

The retrieval mechanism employs exact similarity search to identify the k most semantically similar instances from the evidence set, specifically adapted for evidence-theoretic reasoning rather than direct classification. This distinction is fundamental: while traditional similarity-based classification aggregates neighbor class labels through voting mechanisms, the proposed framework retrieves neighbors to access their pre-computed DS mass functions for subsequent evidence fusion. This approach enables the integration of uncertainty representations rather than mere class predictions, facilitating more nuanced uncertainty-aware decision making.

% The retrieval mechanism employs k-nearest neighbor search adapted for evidence-theoretic reasoning. While conceptually similar to standard k-NN classification, the proposed approach differs fundamentally in its purpose and utilization. Rather than directly using neighbor labels for classification, the framework retrieves neighbors to access their associated Dempster-Shafer mass functions for evidence fusion. This distinction is critical: traditional k-NN aggregates class votes, whereas the proposed method combines uncertainty representations through evidence theory. 

This implementation requires L2 normalization of all feature embeddings to enable cosine similarity computation in the normalized space. This study leverages Facebook AI Similarity Search (FAISS), a library optimized for efficient similarity search and clustering of dense vectors \cite{johnson2017faiss}. FAISS provides multiple indexing strategies designed to handle large-scale vector databases with varying trade-offs between search accuracy, memory consumption, and computational speed. For this implementation, the IndexFlatIP index was selected, which performs exact inner product computation between L2-normalized vectors to achieve precise cosine similarity search. This configuration ensures that retrieved neighbors represent the most semantically similar instances without approximation errors that could compromise evidence quality. The exact search capability is particularly critical for UQ applications where imprecise neighbor selection could propagate uncertainty estimation errors through the evidence fusion process. FAISS IndexFlatIP operates with O(nd) computational complexity for each query, where n represents the evidence set size and d denotes the embedding dimensionality, providing deterministic similarity rankings essential for reproducible uncertainty-aware decisions.

% To implement this, $\ell_2$ normalization was applied to all embeddings to enable cosine similarity computation. The normalized embeddings of the evidence set were indexed using \texttt{FAISS IndexFlatIP} to facilitate efficient exact similarity search.

Mathematically, For each test sample $X^*$, the $k$ most similar instances from the evidence set were retrieved based on cosine similarity, Eq. \ref{Eq:cosine_sim} in the embedding space. 

% The cosine similarity metric quantifies semantic proximity in the normalized embedding space.

% \begin{equation}
%     s(\mathbf{x}, \mathbf{x}_i) = \frac{\mathbf{x} \cdot \mathbf{x}_i}{\|\mathbf{x}\| \, \|\mathbf{x}_i\|}
%     \label{Eq:cosine_sim}
% \end{equation}.

\begin{equation}
    s(\mathbf{x}^*, \mathbf{e}_i) = \frac{\mathbf{x}^* \cdot \mathbf{e}_i}{\|\mathbf{x}^*\| \, \|\mathbf{e}_i\|}
    \label{Eq:cosine_sim}
\end{equation}

\noindent where both vectors are L2-normalized. The FAISS IndexFlatIP returns the k neighbors $\{e_{k_1}, e_{k_2}, \ldots, e_{k_K}\}$ with highest similarity scores.

The next step involves evidence fusion across the retrieved evidence samples through iterative application of Dempster's combination rule. Similar to the test sample processing (section \ref{sec:DS}), M iterations are performed for each evidence $e_k$ to obtain predictions, from which credal intervals are computed at confidence level $\alpha = 0.9$. 

Accordingly, each of the k retrieved neighbors contributes a complete mass function that assigns belief masses to individual classes and ignorance mass to the frame of discernment. Let the frame of discernment be: 

\begin{equation}
    \Omega=\{c_1,\dots,c_C\}.
\end{equation}

\noindent Each retrieved neighbor $e_k$ (where $ k \in \{1, 2, \ldots, K\}$ from K-nearest neighbor set) provides a mass function \(m^{(k)}\) with the following structure:

\begin{itemize}
  \item \(m^{(k)}(\{c_j\})\): belief mass for class \(c_j\in\{c_1,c_2,\dots,c_C\}\),
        where \(m^{(k)}(\{c_j\})=p^{-}_{c_j}(e_k) \) (the lower bound of the credal interval);
  \item \(m^{(k)}(\Omega)\): ignorance mass calculated as:
        \begin{equation}
            m^{(k)}(\Omega)=1-\sum_{j=1}^{C}p^{-}_{c_j}(e_k).
        \end{equation}
\end{itemize}

These K individual mass functions are combined through iterative application of Dempster's combination rule to produce a single consolidated mass function representing the collective evidence. For singleton focal elements \(\{c_j\}\) and any two mass functions
\(m^{(k')}\) and \(m^{(k'')}\) with \(k',k''\in\{1,\dots,K\}\) and \(k'\neq k''\), the combined mass function
\(m^{(k')} \oplus m^{(k'')}\) is computed as in Eq.~\ref{Eq:consolidated_mass}.

\begin{equation}
    \bigl(m^{(k')} \oplus m^{(k'')}\bigr)\!\left(\{c_j\}\right) = \frac{N}{D}
    \label{Eq:consolidated_mass}
\end{equation}
\noindent where
\begin{align}
    N &= m^{(k')}(\{c_j\})\,m^{(k'')}(\{c_j\}) \nonumber \\
      &\quad + m^{(k')}(\{c_j\})\,m^{(k'')}(\Omega) \nonumber \\
      &\quad + m^{(k')}(\Omega)\,m^{(k'')}(\{c_j\}) \\
    D &= 1-\sum_{i=1}^{C}\sum_{\substack{\ell=1\\ \ell\neq i}}^{C} m^{(k')}(\{c_i\})\,m^{(k'')}(\{c_\ell\})
\end{align}

\noindent Conflict resolution is implemented to address numerical instability when \(D \le 1 \times 10^{-10}\), indicating extreme conflict between evidence sources. In such cases, the system assigns a uniform mass distribution:
\[
  m(\{c_j\}) = m(\Omega) = \frac{1}{C+1}, \qquad j=1,\dots,C.
\]

The fusion process proceeds iteratively: starting with the first neighbor's mass function $m^{(k_1)}$, it is combined with $m^{(k_2)}$ to produce $m^{(k_1)} \oplus m^{(k_2)}$. This result is then combined with $m^{(k_3)}$, and so forth, until all $k$ evidences are incorporated:

\begin{equation}
    m_{\text{fused}} = m^{(k_1)} \oplus m^{(k_2)} \oplus \dots \oplus m^{(k_K)}
    \label{Eq:fused_evidence}
\end{equation}

\noindent The final result is a single consolidated mass function \(m_{\text{fused}}\) that
aggregates evidence from all retrieved neighbors, providing class-specific belief
masses \(m_{\text{fused}}(\{c_j\})\) and an overall ignorance mass \(m_{\text{fused}}(\Omega)\)
that reflects the collective ignorance assessment. The class receiving the highest
belief mass from this consolidation represents the evidence-based recommendation,
reflecting the strongest collective support from the neighborhood of similar samples.

\subsection{Uncertainty-aware Decision Framework}
The framework replaces fixed global thresholds with adaptive, sample-specific decision criteria derived from local neighborhood evidence, enabling contextually-aware uncertainty assessment rather than uniform threshold application.

\noindent \textbf{Acceptance Rule}:

For a test instance $x^*$, the individual MCD prediction yields a mass function $m_{x^*}$, while the neighborhood evidence fusion (Eq.~\ref{Eq:fused_evidence}) produces a consolidated mass function $m_{\text{fused}}$.  

\noindent Let the predicted class from the individual assessment be:
\begin{equation}
    \hat y_{\text{ind}}(x^*) = \arg\max_{c \in \{1,\dots,C\}} m_{x^*}(c),
\end{equation}

\noindent and the class supported by the neighborhood evidence fusion be:
\begin{equation}
    \hat y_{\mathcal{N}}(x^*) = \arg\max_{c \in \{1,\dots,C\}} m_{\text{fused}}(c).
\end{equation}

\noindent Define the belief strength of the individual prediction as:
\begin{equation}
    \mathrm{Bel}_{x^*} = m_{x^*}\!\big(\hat y_{\text{ind}}(x^*)\big)
\end{equation}

\noindent and the belief strength of the fused evidences as
\begin{equation}
    \mathrm{Bel}_{\mathcal{N}} = m_{\text{fused}}\!\big(\{\hat y_{\mathcal{N}}(x^*)\}\big).
\end{equation}

\noindent A minimum belief threshold $\tau \in (0,1)$ is enforced. 
The uncertainty-aware decision rule is then:
\begin{equation}
\delta(x^*) =
\begin{cases}
    \text{Certain}, & \text{if } \hat y_{\text{ind}}(x^*) = \hat y_{\mathcal{N}}(x^*) , \\
         &\quad  \text{and} \;\; \hat y_{\text{ind}}(x^*) \neq \Omega, \\
         &\quad  \text{and} \;\; \hat y_{\mathcal{N}}(x^*) \neq \Omega, \\
         &\quad  \text{and} \;\; \mathrm{Bel}_{x^*} \geq \tau, \\
         &\quad  \text{and} \;\; \mathrm{Bel}_{\mathcal{N}} \geq \tau \\[2mm]
    \text{Uncertain}, & \text{otherwise}.
\end{cases}
\end{equation}

\noindent This criterion ensures that: (1) both the individual assessment and the evidences
consensus must agree on a specific class (not ignorance); (2) neither assessment selects ignorance as
the most supported outcome; and (3) both demonstrate sufficient belief strength. The threshold $\tau$ acts as a safety valve, ensuring that only predictions with sufficient evidential support are acted upon automatically. 

The conditions \(\hat{y}_{\text{ind}}(x^{\ast}) \neq \Omega\) and \(\hat{y}_{\mathcal{N}}(x^{\ast}) \neq \Omega\) explicitly assign uncertainty cases where the
highest mass is assigned to ignorance, effectively treating ignorance as an \textit{"I don't know"} class that automatically triggers human review. This prevents the system from making certain decisions when
the evidences fundamentally indicates epistemic uncertainty.

% The threshold $\tau$ prevents acceptance of low-confidence predictions even in cases where consensus exists, thereby improving robustness against unreliable evidences.

\section{Experiments} \label{sec:Experiments}

\subsection{Datasets}
In this study, the proposed framework is evaluated using the CIFAR-10 and CIFAR-100 datasets, both of which are widely recognized as standard benchmarks for image classification tasks. The CIFAR-10 dataset comprises 60,000 color images with a resolution of 32×32 pixels, evenly distributed across 10 broad object categories with 6,000 samples per class.

To complement the insights obtained from CIFAR-10, the CIFAR-100 dataset is employed to evaluate the framework under higher levels of visual complexity and semantic granularity.

Similarly, the CIFAR-100 comprises 60,000 color images at a resolution of 32×32 pixels, uniformly distributed across 100 fine-grained object categories, each consisting of 600 samples. These categories are organized into 20 coarse-grained superclasses, allowing assessment across multiple levels of semantic abstraction. To examine model performance under varying degrees of granularity, experiments are conducted on both the full 100-class configuration and the 20-class semantic category variant. 

% The 100-class configuration allows assessment of the model’s ability to distinguish subtle differences between visually similar categories, while the 20-class variant offers insight into performance when categories are aggregated by semantic similarity.

The CIFAR-10 and CIFAR-100 datasets were chosen for its general-purpose image content, which allows for model evaluation and sample interpretation without requiring domain-specific expertise. This is particularly advantageous for analyzing proximity-based retrieval, where visual assessment of similar samples is critical. 

% Additionally, the CIFAR-100 provides both a fine-grained 100-class setting and a coarse-grained 20-class semantic grouping. This dual structure enables a comprehensive examination of the proposed framework across varying levels of class granularity. The 100-class configuration allows assessment of the model’s ability to distinguish subtle differences between visually similar categories, while the 20-class variant offers insight into performance when categories are aggregated by semantic similarity. Together, these perspectives support a robust analysis of both prediction accuracy and uncertainty behavior across abstraction levels.

\subsection{Configuration}
To mitigate the computational burden of training convolutional neural networks (CNNs) from scratch and to leverage the benefits of previously acquired knowledge, this study adopts a transfer learning approach. Pretrained models, originally trained on large-scale datasets such as ImageNet, are utilized as fixed feature extractors by discarding their classification layers and retaining the convolutional backbone in a frozen state. This enables the extraction of high-level, domain-agnostic features from the target dataset without additional training overhead. In this study, two pretrained models with contrasting architectural paradigms are employed: BiT, a large-scale model based on convolutional residual networks, and ViT, a transformer-based vision architecture. Although the native output dimensionalities of these models differ due to their structural differences, a global max pooling operation is applied to the final feature representations to project them into a fixed 256-dimensional embedding space. This transformation ensures consistency in the downstream pipeline. The inclusion of both a convolutional residual network-based and a transformer-based architecture allows for an investigation into how backbone design influences classification performance and uncertainty estimation.

\subsection{Evaluation}
In order to capture the full scope of the method’s performance, the evaluation procedure is organized into two complementary parts that collectively ensure a complete appraisal:

\begin{itemize}
    \item \textbf{Predictive Performance}: Standard performance evaluation employs established classification metrics including F1-score, accuracy, and area under the curve (AUC), complemented by calibration metrics to assess prediction reliability. Expected Calibration Error (ECE) quantifies the discrepancy between predicted confidence and actual accuracy across probability bins formulated by Eq. \ref{Eq:ECE}:
    \begin{equation}
        \mathrm{ECE} = \frac{|b_m|}{N} \sum_{m=1}^{M}  \, \big| \mathrm{acc}(b_m) - \mathrm{conf}(b_m) \big|
    \label{Eq:ECE}
    \end{equation}
    where $B_m$ represents the $m$-th bin, $|B_m|$ denotes the bin size, $n$ is the total sample count, $\text{acc}(B_m)$ is the accuracy within bin $m$ as calculated Eq. \ref{Eq:ECE_acc}, and $\text{conf}(B_m)$ is the average confidence in that bin calculated by Eq. \ref{Eq:ECE_conf}. 
    
    \begin{equation}
        \mathrm{acc}(b_m) = \frac{1}{|b_m|} \sum_{i \in B_m} \mathbf{1}(\hat{y}_i = y_i).
    \label{Eq:ECE_acc}
    \end{equation}

    \begin{equation}
        \mathrm{conf}(b_m) = \frac{1}{|b_m|} \sum_{i \in B_m} p_i.
    \label{Eq:ECE_conf}
    \end{equation}
    where $y_i$ denotes the ground-truth label of sample $i$, $\hat{y}_i$ is the predicted label for sample $i$, and $p_i$ represents the estimated confidence score(i.e., the probability assigned to $\hat{y}_i$). 

    Additionally, the Brier Score provides an alternative calibration assessment by measuring the mean squared difference between predicted probabilities and actual outcomes Eq. \ref{Eq:BS}: 
    \begin{equation}
        \text{BS} = \frac{1}{N} \sum_{i=1}^{N}\sum_{c=1}^{C} (p_{i,c} - y_{i,c})^2
        \label{Eq:BS}
    \end{equation}

    where $p_{i,c}$ represents the predicted probability for class $c$ of sample $i$, $y_{i,c}$ is the true label indicator, $C$ is the number of classes, and $N$ denotes the total number of predictions. A smaller Brier Score reflects superior predictive performance, since the Brier Score is minimized when predictions are of higher quality, as it jointly penalizes lack of calibration and unnecessary uncertainty \cite{gharoun2025leveraging}.

    \item \textbf{Uncertainty-aware Decisions Evaluation}: The uncertainty-aware decision framework categorizes predictions into four distinct outcomes based on the intersection of prediction correctness and certainty assessment,as introduced by \cite{asgharnezhad2022objective}: (\textbf{\romannum{1}}) \textit{True Certainty (TC)} refers to cases in which the model produces a correct prediction accompanied by low uncertainty, (\textbf{\romannum{2}}) \textit{True Uncertainty (TU)} captures instances where the prediction is incorrect and the model expresses high uncertainty, demonstrating awareness of its own limitations, (\textbf{\romannum{3}}) \textit{False Uncertainty (FU)} describes correct predictions that are associated with high epistemic uncertainty, leading to the unnecessary review of correct outputs, (\textbf{\romannum{4}}) \textit{False Certainty (FC)} represents incorrect predictions made under low epistemic uncertainty, reflecting unwarranted model certainty. 

    \begin{figure}[H]
          \centering
          \captionsetup[subfloat]{font=tiny}
          {\includegraphics[width=\columnwidth]{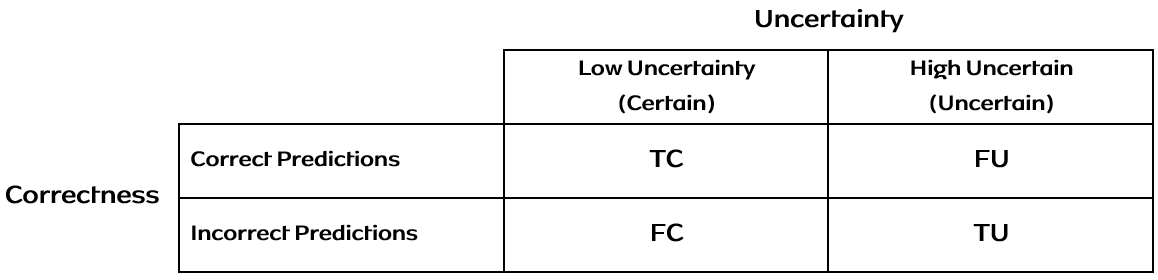}}\par
          \caption{Uncertainty-Informed Outcome Typology}
          \label{fig:U_conf_mat}
    \end{figure}
 
    Dividing the model's predictions into four distinct categories, as shown in Figure \ref{fig:U_conf_mat}, facilitates the derivation of evaluation metrics that parallel those of the conventional confusion matrix, but are specifically tailored to assess uncertainty-aware decisions. In this work, the following uncertainty-aware performance measures are employed:

    \begin{itemize}
        \item Uncertainty Accuracy (UAcc): quantifies the proportion of predictions for which both the predicted label and the associated uncertainty status align with their true counterparts. It reflects the model’s overall reliability in uncertainty-aware decision-making and is computed as defined in Eq. \ref{Eq:UACC}.
        
        \ref{Eq:UACC}.
        \begin{equation}
            UAcc= \frac{TU+TC}{TU+TC+FU+FC}
        \label{Eq:UACC}
        \end{equation}
        
        \item Uncertainty True Positive Rate (UTPR): evaluates the model’s effectiveness in recognizing predictions that warrant certainty, specifically among those that are truly correct. It is calculated using Eq. \ref{Eq:UTPR}. A higher value reflects the model’s improved capacity to assign certain decisions to accurate predictions.
        
        \begin{equation}
            UTPR= \frac{TC}{TC+FU}
        \label{Eq:UTPR}
        \end{equation}

        \item Uncertainty False Positive Rate (UFPR): captures the proportion of incorrect predictions that are mistakenly treated as certain, relative to all cases where uncertainty would be expected (i.e., incorrect predictions). It is defined by Eq. \ref{Eq:UFPR}. A lower value suggests improved model safety, as it reflects a reduced tendency to convey unjustified certainty in incorrect outcomes.
        
        \begin{equation}
            UFPR= \frac{FC}{FC+TU}
        \label{Eq:UFPR}
        \end{equation}
        
        \item Uncertainty G-Mean (UG-Mean): provides a balanced measure that simultaneously considers both the model's ability to confidently handle correct predictions (UTPR) and its capacity to appropriately flag incorrect predictions as uncertain (Uncertainty Specificity), calculated by Eq. \ref{Eq:UGMEAN}. Uncertainty Specificity can be obtained by $1-UFPR$. This metric is particularly valuable for interpreting results as it captures the trade-off between operational efficiency and safety in uncertainty-aware systems, with higher values indicating superior overall uncertainty quantification performance.
        \begin{equation}
            UG-Mean= \sqrt{\text{UTPR} \times (1 - \text{UFPR})}
        \label{Eq:UGMEAN}
        \end{equation}

    \end{itemize}

\end{itemize}

\section{Results and Discussion} \label{sec:Results}
This section reports the experimental results obtained using the proposed method. To contextualize the performance results, the model selection and optimization process is first described. To optimize the neural network architecture, Keras Tuner was employed to explore a predefined hyperparameter space. The hyperparameter search space encompassed network depth (1-4 hidden layers), layer widths (32-512 neurons), and learning rates on a logarithmic scale ($10^{-4}$ to $10^{-2}$). Each hidden layer incorporated batch normalization followed by ReLU activation and dropout regularization. The output layer utilized softmax activation for multi-class probability estimation. The tuning process was guided by the objective of maximizing model accuracy, and the best-performing configuration was selected for further evaluation. The finalized architecture was then tested over 20 independent runs. In each run, the dataset was randomly partitioned into three subsets: a training set, an evidence set, and a test set, with respective proportions of 70\%, 15\%, and 15\%. This repeated sampling strategy enables the model to encounter a variety of data distributions, thereby providing a more comprehensive and robust assessment of its generalization capability. The best results obtained across these runs are presented in Table~\ref{tab:accuracy_result}.

\begin{table*}
\centering
\caption{Classification results obtained over 20 independent runs on CIFAR10 \& both the fine-grained (100 classes) and coarse-grained (20 superclasses) of CIFAR-100 settings.}
\label{tab:accuracy_result}
\resizebox{0.8\textwidth}{!}{%
\begin{tblr}{
  row{1} = {c},
  hline{1-2,8} = {-}{},
}
Dataset & Backbone & F1 score & Accuracy & AUC & ECE & Brier  Score\\
CIFAR10 & BiT & 90.9862 ± 0.1971 & 91.0167 ± 0.1939 & 95.0093 ± 0.1077 & 5.2134 ± 0.2429 & 14.4887 ± 0.1799\\
CIFAR10 & ViT & 95.7722 ± 0.2431 & 95.7722 ± 0.2454 & 97.6512 ± 0.1363 & 0.6938 ± 0.2783 & 6.3478 ± 0.4008\\
CIFAR100 (Superclasses) & BiT & 79.3675 ± 0.2676 & 79.41 ± 0.2619 & 89.1632 ± 0.1378 & 8.9479 ± 0.5852 & 30.7445 ± 0.2651\\
CIFAR100~(Superclasses) & ViT & 83.328 ± 0.2305 & 83.3389 ± 0.2294 & 91.2094 ± 0.1266 & 16.9443 ± 1.4456 & 27.788 ± 0.2242\\
CIFAR100 & BiT & 69.4701 ± 0.3717 & 69.6411 ± 0.3483 & 84.6672 ± 0.1759 & 15.279 ± 0.482 & 44.2044 ± 0.2524\\
CIFAR100 & ViT & 78.306 ± 0.2682 & 78.427 ± 0.2429 & 89.1046 ± 0.1227 & 28.9157 ± 0.3033 & 40.1565 ± 0.2289
\end{tblr}
}
\end{table*}

The baseline classification results demonstrate a clear performance hierarchy, with ViT consistently outperforming BiT across all experimental settings. On CIFAR-10, ViT achieved 95.77\% accuracy compared to BiT's 91.02\%, representing a substantial 4.8 percentage point improvement. This performance gap widened further on the more challenging CIFAR-100 fine-grained classification task, where ViT reached 78.43\% accuracy while BiT achieved only 69.64\%, an 8.8 percentage point difference. The CIFAR-100 superclass task showed intermediate results, with ViT at 83.34\% and BiT at 79.41\%. The dataset complexity analysis reveals expected performance degradation as classification difficulty increases. The low standard deviations across all metrics confirm the reproducibility and statistical robustness of these baseline results, with typical variations of only 0.1-0.5\% across 20 independent runs.

These findings highlight a fundamental trade-off between classification accuracy and confidence calibration. While ViT demonstrates superior accuracy across all datasets, it exhibits significantly worse calibration compared to BiT, especially on complex classification tasks. On CIFAR-10, ViT achieves both excellent accuracy (95.77\%) and outstanding calibration (0.69\% ECE), suggesting optimal performance on this relatively simple 10-class problem. However, this calibration quality deteriorates dramatically on CIFAR-100, where ViT reaches 28.92\% ECE despite maintaining higher accuracy than BiT. In contrast, BiT demonstrates more consistent calibration behavior across datasets, maintaining ECE values between 5-15\% regardless of task complexity, though at the cost of lower classification accuracy. This pattern suggests that BiT's residual architecture produces more reliable confidence estimates. These findings provide additional empirical support for the observations of \cite{guo2017calibration} regarding the inherent miscalibration of modern neural networks.

These baseline results establish the foundation for the subsequent uncertainty quantification framework. It is important to note that ideal uncertainty-aware decision making is characterized by the model being certain when correct, and uncertain when likely to be incorrect. Under such conditions, users can reliably trust predictions labeled as certain, while directing attention to those flagged as uncertain. In pursuit of this ideal, effective uncertainty-aware systems should strive to minimize FC while maximizing TC. Additionally, FU should be minimized to avoid excessive and unnecessary reviews by users. The evaluation framework presented in the previous section is designed to assess how closely a system approaches this ideal behavior across these key dimensions.

\begin{table*}[t]
\centering
\caption{Uncertainty-aware performance summary of the proposed method on CIFAR-10 \& 100.}
\label{tab:PEN_result}
\begin{threeparttable}
\resizebox{0.9\textwidth}{!}{%
\begin{tblr}{
  row{1} = {c},
  hline{1-2,8} = {-}{},
}
Dataset & Backbone & Params* & UAcc & UTPR & UFPR & UG-Mean & ~TC & ~FC & ~TU & ~FU\\
CIFAR10 & BiT & (3, 0.3) & 81.3211 ± 0.7429 & 80.7287 ± 0.9018 & 12.7597 ± 1.4261 & 83.9146 ± 0.5056 & 6,606 & 104 & 713 & 1,577\\
CIFAR10 & ViT & (3, 0.6) & 89.4467 ± 0.6962 & 89.4241 ± 0.7556 & 10.0426 ± 1.398 & 89.6854 ± 0.602 & 7,709 & 38 & 341 & 912\\
CIFAR100 (Superclasses) & BiT & (3, 0.1) & 75.5528 ± 0.4047 & 74.352 ± 0.4559 & 20.002 ± 0.6699 & 77.1225 ± 0.425 & 5,269 & 383 & 1,530 & 1,818\\
CIFAR100 (Superclasses) & ViT & (3, 0.15) & 76.68 ± 0.3531 & 74.0296 ± 0.4766 & 10.5034 ± 0.8974 & 81.3945 ± 0.378 & 5,521 & 162 & 1,380 & 1,937\\
CIFAR100 & BiT & (3, 0.1) & 65.4772 ± 0.4582 & 53.3891 ± 0.6954 & 6.6654 ± 0.5454 & 70.5882 ± 0.3933 & 3,351 & 182 & 2,542 & 2,926\\
CIFAR100 & ViT & (3, 0.1) & 70.9628 ± 0.4174 & 64.7301 ± 0.4681 & 6.2828 ± 0.4042 & 77.8859 ± 0.3295 & 4,573 & 122 & 1,814 & 2,492
\end{tblr}
}
\begin{tablenotes}
\small
\item * Parameters of the proposed method: ($k$ (evidence set size), Belief Threshold). Values for TC, FC, TU, and FU represent sample counts averaged over 20 runs and rounded to the nearest integer.
\end{tablenotes}
\end{threeparttable}
\end{table*}

Table \ref{tab:PEN_result} reports the uncertainty-aware evaluation results for the proposed method. In these experiments, the number of neighbors was fixed at $k=3$, and the belief threshold was determined via grid search over the range $[0.1,0.6]$, with the selection criterion being the maximization of UG-Mean while minimizing FC. Table \ref{tab:Threshold_result} presents the performance of a baseline entropy-thresholding approach, where PE thresholds were varied between 0.05 and 0.75. For this baseline, the threshold yielding the highest UG-Mean with the lowest FC is shown in boldface. 

The following comparison contrasts the results in Table \ref{tab:PEN_result} with the best-performing configuration (boldface) from Table \ref{tab:Threshold_result}.

% Consistent with this principle, the evaluation metrics defined in the previous section are designed to minimize FC (confidently incorrect predictions, which are most harmful) while maximizing TC (truly confident correct predictions). Additionally, FU is also minimized, since high FU values imply excessive and unnecessary reviews by users.

Across all datasets, the proposed method shows favorable trends in TC and FC, while maintaining comparable or lower TU. For CIFAR-10 with the BiT backbone, the proposed method not only improves UG-Mean (83.91\% vs. 82.49\%) but also lowers FC from 121 to 104, meaning fewer certain wrong decisions are made. TC remains comparable between the two approaches, demonstrating that the gain in reliability is not achieved at the expense of correct certain outcomes. Similarly, for CIFAR-10 with the ViT backbone, the proposed method enhances UG-Mean (89.69\% vs. 88.50\%) while reducing FC (38 vs. 44) and slightly improving TC, thereby delivering a better balance of confident and correct decisions.

For CIFAR-100 Superclasses, improvements are also evident. With the BiT backbone, the proposed method achieves a higher UG-Mean (77.12\% vs. 74.67\%) while lowering FC from 397 to 383. TC is slightly higher as well, showing that the method increases the number of reliable certain predictions. With the ViT backbone, the benefits are more marked: the proposed method raises UG-Mean from 78.71\% to 81.39\%, decreases FC from 193 to 162, and increases TC, highlighting its capacity to enhance reliability in broader categorical structures.

In the CIFAR-100 fine-grained setting, a nuanced picture emerges. With the BiT backbone, entropy achieves a marginally higher UG-Mean (72.10\% vs. 70.59\%), but this comes at the cost of a dramatic increase in FC (599 vs. 182). In contrast, the proposed method keeps FC much lower, reflecting fewer certain wrong predictions, albeit with a slight reduction in TC. With the ViT backbone, the advantage of the proposed approach is again clear: UG-Mean rises from 72.89\% to 77.89\%, TC is higher, and FC drops from 147 to 122, confirming its ability to deliver more trustworthy decisions. TU values remain comparable across methods, suggesting that improvements are achieved without forcing excessive deferrals.

\begin{table*}[t]
\centering
\caption{Uncertainty-aware performance summary obtained by applying a threshold to prediction entropy.}
\label{tab:Threshold_result}
\resizebox{0.9\textwidth}{!}{%
\begin{tblr}{
  row{1} = {c},
  cell{2}{2} = {c},
  cell{2}{3} = {c},
  cell{3}{2} = {c},
  cell{3}{3} = {c},
  cell{4}{2} = {c},
  cell{4}{3} = {c},
  cell{5}{2} = {c},
  cell{5}{3} = {c},
  cell{6}{2} = {c},
  cell{6}{3} = {c},
  cell{7}{2} = {c},
  cell{7}{3} = {c},
  cell{8}{2} = {c},
  cell{8}{3} = {c},
  cell{9}{2} = {c},
  cell{9}{3} = {c},
  cell{10}{2} = {c},
  cell{10}{3} = {c},
  cell{11}{2} = {c},
  cell{11}{3} = {c},
  cell{12}{2} = {c},
  cell{12}{3} = {c},
  cell{13}{2} = {c},
  cell{13}{3} = {c},
  cell{14}{2} = {c},
  cell{14}{3} = {c},
  cell{15}{2} = {c},
  cell{15}{3} = {c},
  cell{16}{2} = {c},
  cell{16}{3} = {c},
  cell{17}{2} = {c},
  cell{17}{3} = {c},
  cell{18}{2} = {c},
  cell{18}{3} = {c},
  cell{19}{2} = {c},
  cell{19}{3} = {c},
  cell{20}{2} = {c},
  cell{20}{3} = {c},
  cell{21}{2} = {c},
  cell{21}{3} = {c},
  cell{22}{2} = {c},
  cell{22}{3} = {c},
  cell{23}{2} = {c},
  cell{23}{3} = {c},
  cell{24}{2} = {c},
  cell{24}{3} = {c},
  cell{25}{2} = {c},
  cell{25}{3} = {c},
  cell{26}{2} = {c},
  cell{26}{3} = {c},
  cell{27}{2} = {c},
  cell{27}{3} = {c},
  cell{28}{2} = {c},
  cell{28}{3} = {c},
  cell{29}{2} = {c},
  cell{29}{3} = {c},
  cell{30}{2} = {c},
  cell{30}{3} = {c},
  cell{31}{2} = {c},
  cell{31}{3} = {c},
  cell{32}{2} = {c},
  cell{32}{3} = {c},
  cell{33}{2} = {c},
  cell{33}{3} = {c},
  cell{34}{2} = {c},
  cell{34}{3} = {c},
  cell{35}{2} = {c},
  cell{35}{3} = {c},
  cell{36}{2} = {c},
  cell{36}{3} = {c},
  cell{37}{2} = {c},
  cell{37}{3} = {c},
  cell{38}{2} = {c},
  cell{38}{3} = {c},
  cell{39}{2} = {c},
  cell{39}{3} = {c},
  cell{40}{2} = {c},
  cell{40}{3} = {c},
  cell{41}{2} = {c},
  cell{41}{3} = {c},
  cell{42}{2} = {c},
  cell{42}{3} = {c},
  cell{43}{2} = {c},
  cell{43}{3} = {c},
  hline{1-2,44} = {-}{},
}
Dataset & Backbone & {Confidence\\Threshold} & UAcc & UTPR & UFPR & UG-Mean & TC & FC & TU & FU\\
CIFAR10 & BiT & 0.05 & 10.5048 ± 0.5097 & 1.6637 ± 0.5076 & 0.0 ± 0.0 & 12.7467 ± 1.974 & 136 & - & 809 & 8,055\\
CIFAR10 & BiT & 0.1 & 24.1256 ± 1.6718 & 16.6351 ± 1.804 & 0.0448 ± 0.0736 & 40.7172 ± 2.1913 & 1,362 & - & 809 & 6,828\\
CIFAR10 & BiT & 0.2 & 56.2215 ± 1.2653 & 52.0386 ± 1.4073 & 1.4351 ± 0.3938 & 71.6107 ± 0.9146 & 4,262 & 12 & 798 & 3,928\\
CIFAR10 & BiT & 0.3 & 71.6622 ± 0.7509 & 69.4235 ± 0.8682 & 5.6762 ± 0.7931 & 80.9179 ± 0.4015 & 5,686 & 46 & 763 & 2,504\\
\textbf{CIFAR10} & \textbf{BiT} & \textbf{0.4} & \textbf{80.4822 ± 0.5643} & \textbf{80.0326 ± 0.7074} & \textbf{14.9663 ± 1.5041} & \textbf{82.4891 ± 0.5485} & \textbf{6,555} & \textbf{121} & \textbf{688} & \textbf{1,636}\\
CIFAR10 & BiT & 0.5 & 86.1015 ± 0.3746 & 87.7266 ± 0.5133 & 30.3528 ± 2.0588 & 78.1549 ± 1.0225 & 7,186 & 246 & 564 & 1,005\\
CIFAR10 & BiT & 0.75 & 91.3493 ± 0.1924 & 98.24 ± 0.2107 & 78.3984 ± 2.1822 & 46.0082 ± 2.2557 & 8,047 & 634 & 175 & 144\\
CIFAR10 & ViT & 0.05 & 51.5378 ± 8.0255 & 49.4686 ± 8.387 & 1.527 ± 0.5911 & 69.5545 ± 5.4659 & 4,264 & 6 & 374 & 4,356\\
CIFAR10 & ViT & 0.1 & 77.4967 ± 3.1206 & 76.6938 ± 3.3204 & 4.3887 ± 1.0965 & 85.6019 ± 1.4887 & 6,611 & 7 & 364 & 2,009\\
\textbf{CIFAR10} & \textbf{ViT} & \textbf{0.2} & \textbf{88.5748 ± 1.3621} & \textbf{88.5764 ± 1.4855} & \textbf{11.5603 ± 1.3712} & \textbf{88.4978 ± 0.5041} & \textbf{7,635} & \textbf{44} & \textbf{336} & \textbf{984}\\
CIFAR10 & ViT & 0.3 & 92.3926 ± 0.8879 & 92.9645 ± 0.9624 & 20.6538 ± 1.8816 & 85.8756 ± 0.8423 & 8,014 & 78 & 302 & 606\\
CIFAR10 & ViT & 0.4 & 94.4611 ± 0.5373 & 95.8254 ± 0.5605 & 36.5394 ± 2.6284 & 77.9621 ± 1.5198 & 8,260 & 139 & 241 & 360\\
CIFAR10 & ViT & 0.5 & 95.4956 ± 0.4386 & 97.4984 ± 0.4222 & 49.9913 ± 2.2986 & 69.8062 ± 1.5163 & 8,404 & 190 & 190 & 216\\
CIFAR10 & ViT & 0.75 & 96.0944 ± 0.2605 & 99.649 ± 0.14 & 84.5764 ± 2.5688 & 39.0758 ± 3.1288 & 8,590 & 321 & 59 & 30\\
CIFAR100 (Superclasses) & BiT & 0.05 & 28.1785 ± 0.5318 & 9.552 ± 0.6195 & 0.0432 ± 0.043 & 30.8834 ± 1.0001 & 683 & 1 & 1,854 & 6,463\\
CIFAR100 (Superclasses) & BiT & 0.1 & 36.4981 ± 0.5553 & 20.0757 ± 0.681 & 0.2175 ± 0.084 & 44.7506 ± 0.7599 & 1,435 & 4 & 1,850 & 5,711\\
CIFAR100 (Superclasses) & BiT & 0.2 & 51.18 ± 0.6423 & 38.9857 ± 0.8228 & 1.8293 ± 0.2902 & 61.8612 ± 0.6565 & 2,786 & 34 & 1,820 & 4,360\\
CIFAR100 (Superclasses) & BiT & 0.3 & 63.5007 ± 0.5751 & 56.2189 ± 0.7822 & 8.4387 ± 0.6302 & 71.7431 ± 0.4736 & 4,017 & 156 & 1,698 & 3,128\\
\textbf{CIFAR100 (Superclasses)} & \textbf{BiT} & \textbf{0.4} & \textbf{72.5107 ± 0.4206} & \textbf{70.9276 ± 0.5888} & \textbf{21.3906 ± 0.7922} & \textbf{74.6676 ± 0.37} & \textbf{5,068} & \textbf{397} & \textbf{1,458} & \textbf{2,077}\\
CIFAR100 (Superclasses) & BiT & 0.5 & 78.3885 ± 0.4259 & 82.6437 ± 0.5036 & 38.0149 ± 1.0096 & 71.5699 ± 0.589 & 5,906 & 705 & 1,149 & 1,240\\
CIFAR100 (Superclasses) & BiT & 0.75 & 80.9344 ± 0.2983 & 98.1994 ± 0.1946 & 85.5959 ± 0.7938 & 37.5947 ± 1.0166 & 7,017 & 1,587 & 267 & 129\\
CIFAR100 (Superclasses) & ViT & 0.05 & 16.6778 ± 0.2312 & 0.0231 ± 0.0185 & 0.0 ± 0.0 & 1.3074 ± 0.7757 & 2 & - & 1,499 & 7,499\\
CIFAR100 (Superclasses) & ViT & 0.1 & 17.6711 ± 0.3156 & 1.2149 ± 0.2808 & 0.0 ± 0.0 & 10.9451 ± 1.3037 & 91 & - & 1,499 & 7,410\\
CIFAR100 (Superclasses) & ViT & 0.2 & 30.0189 ± 0.7286 & 16.0872 ± 0.8646 & 0.2825 ± 0.0784 & 40.0372 ± 1.0815 & 1,207 & 4 & 1,495 & 6,294\\
CIFAR100 (Superclasses) & ViT & 0.3 & 48.0426 ± 0.8404 & 37.9159 ± 0.9964 & 1.2936 ± 0.2566 & 61.1709 ± 0.7987 & 2,844 & 19 & 1,480 & 4,657\\
CIFAR100 (Superclasses) & ViT & 0.4 & 63.1163 ± 0.7956 & 56.7218 ± 0.977 & 4.8914 ± 0.4552 & 73.4452 ± 0.5744 & 4,255 & 73 & 1,426 & 3,246\\
\textbf{CIFAR100 (Superclasses)} & \textbf{ViT} & \textbf{0.5} & \textbf{73.78 ± 0.6521} & \textbf{71.1144 ± 0.8092} & \textbf{12.8801 ± 0.844} & \textbf{78.7085 ± 0.491} & \textbf{5,334} & \textbf{193} & \textbf{1,306} & \textbf{2,167}\\
CIFAR100 (Superclasses) & ViT & 0.75 & 85.3548 ± 0.2703 & 93.5864 ± 0.3677 & 55.8245 ± 1.3095 & 64.2901 ± 0.9 & 7,020 & 837 & 662 & 481\\
CIFAR100 & BiT & 0.05 & 34.6026 ± 0.515 & 6.0596 ± 0.4987 & 0.0317 ± 0.0263 & 24.5913 ± 1.0143 & 380 & 1 & 2,735 & 5,885\\
CIFAR100 & BiT & 0.1 & 40.4344 ± 0.6928 & 14.5189 ± 0.8089 & 0.2155 ± 0.0751 & 38.0478 ± 1.058 & 909 & 6 & 2,730 & 5,355\\
CIFAR100 & BiT & 0.2 & 52.3822 ± 0.7027 & 32.531 ± 0.9342 & 2.1553 ± 0.2533 & 56.4116 ± 0.7839 & 2,038 & 59 & 2,677 & 4,227\\
CIFAR100 & BiT & 0.3 & 62.8237 ± 0.522 & 50.6926 ± 0.7569 & 9.392 ± 0.5741 & 67.7699 ± 0.4673 & 3,176 & 257 & 2,479 & 3,089\\
\textbf{CIFAR100} & \textbf{BiT} & \textbf{0.4} & \textbf{70.0763 ± 0.5042} & \textbf{66.5761 ± 0.653} & \textbf{21.9057 ± 0.7352} & \textbf{72.1038 ± 0.4819} & \textbf{4,171} & \textbf{599} & \textbf{2,136} & \textbf{2,094}\\
CIFAR100 & BiT & 0.5 & 74.39 ± 0.417 & 79.498 ± 0.6136 & 37.3031 ± 1.0007 & 70.5958 ± 0.5127 & 4,980 & 1,021 & 1,715 & 1,284\\
CIFAR100 & BiT & 0.75 & 72.8715 ± 0.3363 & 98.0155 ± 0.2155 & 84.7083 ± 0.9223 & 38.6961 ± 1.1484 & 6,140 & 2,317 & 418 & 124\\
CIFAR100 & ViT & 0.05 & 21.5504 ± 0.2043 & 0.0019 ± 0.006 & 0.0 ± 0.0 & 0.1354 ± 0.4128 & - & - & 1,939 & 7,060\\
CIFAR100 & ViT & 0.1 & 21.7907 ± 0.2292 & 0.3083 ± 0.0915 & 0.0 ± 0.0 & 5.4901 ± 0.8316 & 22 & - & 1,939 & 7,039\\
CIFAR100 & ViT & 0.2 & 28.6833 ± 0.5459 & 9.1076 ± 0.6207 & 0.0482 ± 0.0461 & 30.1541 ± 1.0212 & 643 & 1 & 1,938 & 6,418\\
CIFAR100 & ViT & 0.3 & 42.1996 ± 0.5742 & 26.4508 ± 0.7351 & 0.4656 ± 0.1017 & 51.3054 ± 0.7156 & 1,868 & 9 & 1,930 & 5,193\\
CIFAR100 & ViT & 0.4 & 54.5793 ± 0.5012 & 42.7934 ± 0.6381 & 2.5127 ± 0.2782 & 64.5875 ± 0.4666 & 3,021 & 49 & 1,891 & 4,039\\
\textbf{CIFAR100} & \textbf{ViT} & \textbf{0.5} & \textbf{65.0152 ± 0.5062} & \textbf{57.4866 ± 0.6686} & \textbf{7.5773 ± 0.6192} & \textbf{72.8887 ± 0.4171} & \textbf{4,059} & \textbf{147} & \textbf{1,792} & \textbf{3,002}\\
CIFAR100 & ViT & 0.75 & 81.3519 ± 0.3481 & 86.8834 ± 0.4598 & 38.7858 ± 1.1634 & 72.9238 ± 0.654 & 6,134 & 752 & 1,187 & 926
\end{tblr}
}
\begin{tablenotes}
\small
\item Note: TC, FC, TU, and FU values represent sample counts averaged over 20 runs and rounded to the nearest integer.
\end{tablenotes}
\end{table*}

The closer looks at other thresholds results in Table \ref{tab:Threshold_result} demonstrate systematic deficiencies that underscore the limitations of simple threshold-based uncertainty-aware decision making approach. At conservative threshold settings (e.g., 0.05–0.2), entropy methods almost eliminate FC, which might appear favorable compared to the proposed method. However, this comes at a severe cost: UTPR and TC collapse, UG-Mean remains low, and FU values become excessively high, indicating that most predictions are deferred to review. Such behavior renders the model operationally inefficient despite superficially low FC.

At mid-range thresholds (around 0.3–0.4), entropy baselines achieve their strongest performance, with UG-Mean values approaching those of the proposed method. For example, in CIFAR-10 (BiT), a 0.3 threshold produces UG-Mean of 80.9\%, close to the 83.9\% achieved by proposed framework with lower FC. Similar behavior observed for CIFAR-100 Superclasses (BiT at threshold 0.3 and ViT at threshold 0.4) where lower FC is achieved with the rises in the UG-Mean. Same pattern is observed for CIFAR-100 fine-grained (BiT) at threshold 0.2 and CIFAR-100 fine-grained (ViT) at threshold 0.4. However, this comes at the expense of inflated FU and lower TC, meaning that many correct predictions are sacrificed to achieve modest FC control. For instance, CIFAR-10 (BiT) at threshold 0.3 FU is $\approx2,500$, around 50\% higher than the $\approx1,500$ FU of the proposed method. CIFAR-100 Superclasses (BiT) FU at threshold 0.3 $\approx3100$, approximately 70\% greater than the $\approx1800$ FU of the proposed method reflecting reduced operational efficiency caused by excessive deferrals.

Finally, liberal threshold settings create dangerous overconfidence patterns that compromise system safety. At high thresholds (0.5–0.75), entropy methods increase TC and overall accuracy, but this gain is undermined by sharp increases in FC, sometimes exceeding 2–6 times the values observed under the proposed method. Such certain wrong predictions are the most dangerous outcome in safety-critical settings. 

\begin{figure*}[]
\centering
\subcaptionbox{\scriptsize FC decision, Predicted label: Fox, True label: Lion}[0.2\textwidth]{%
  \includegraphics[width=0.2\textwidth]{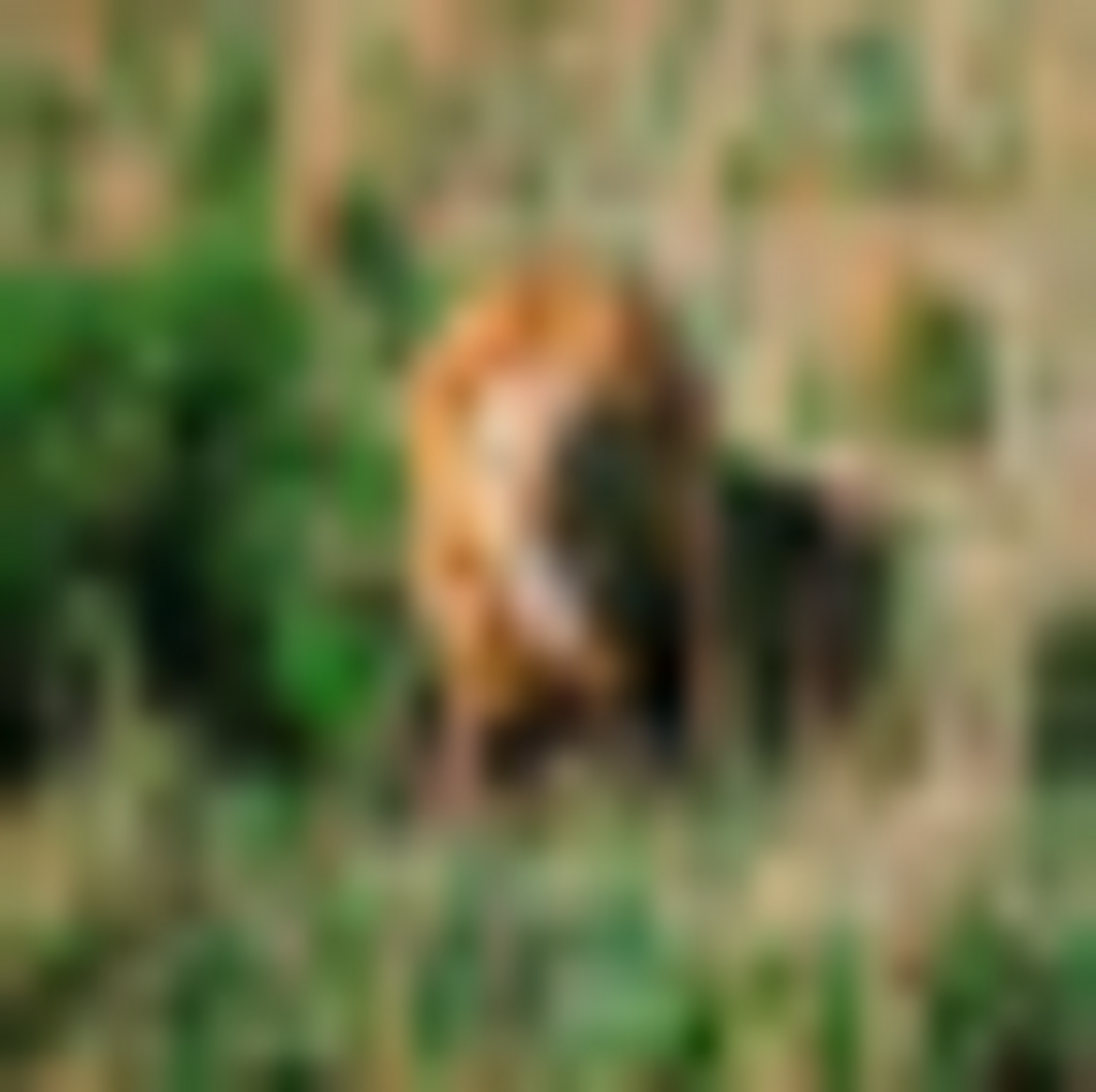}}
\hfill
\subcaptionbox{\scriptsize Evidence 1: True label: Fox, Dist: 0.631}[0.2\textwidth]{%
  \includegraphics[width=0.2\textwidth]{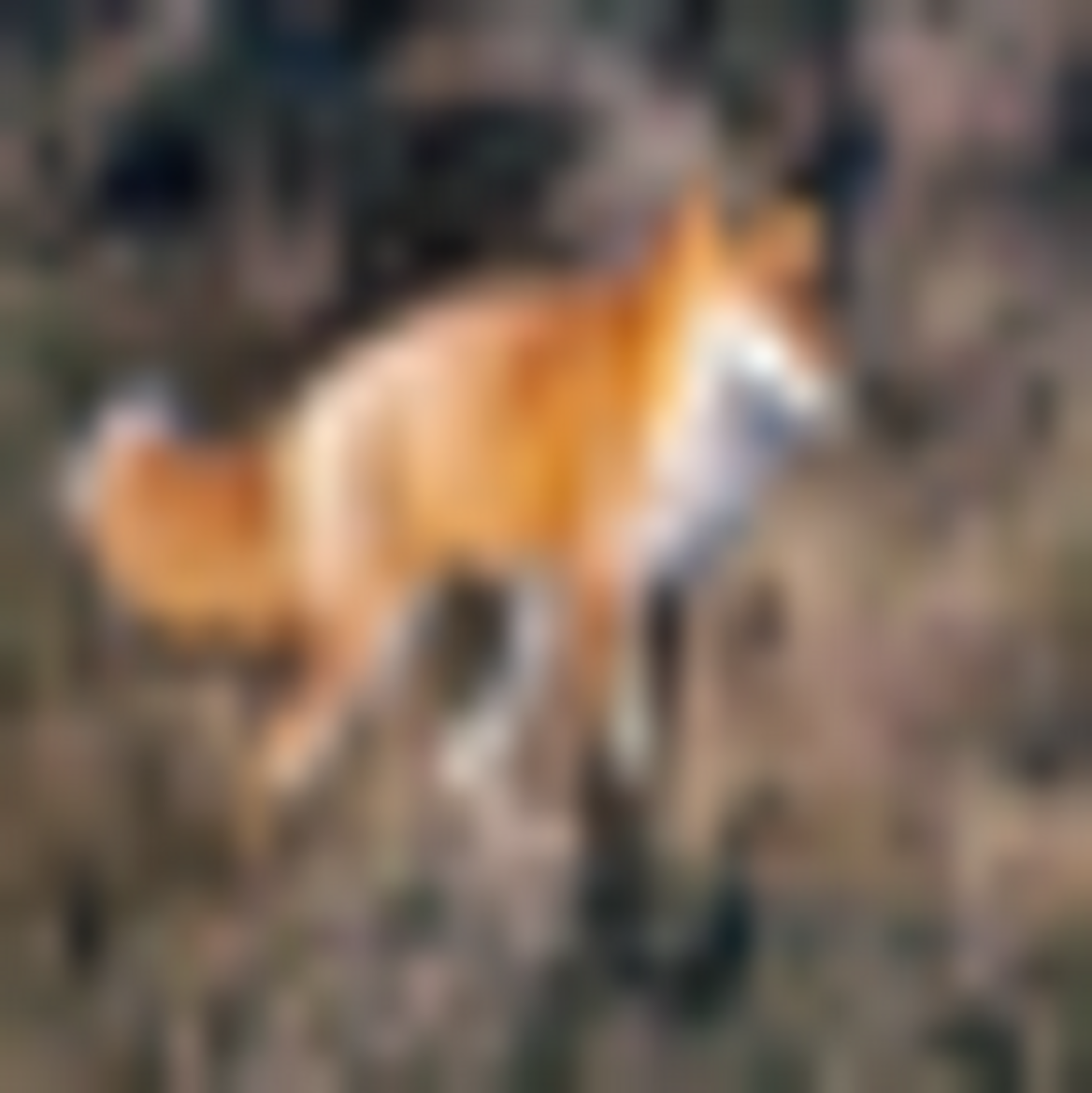}}
\hfill
\subcaptionbox{\scriptsize  Evidence 2: True label: Fox, Dist:0659}[0.2\textwidth]{%
  \includegraphics[width=0.2\textwidth]{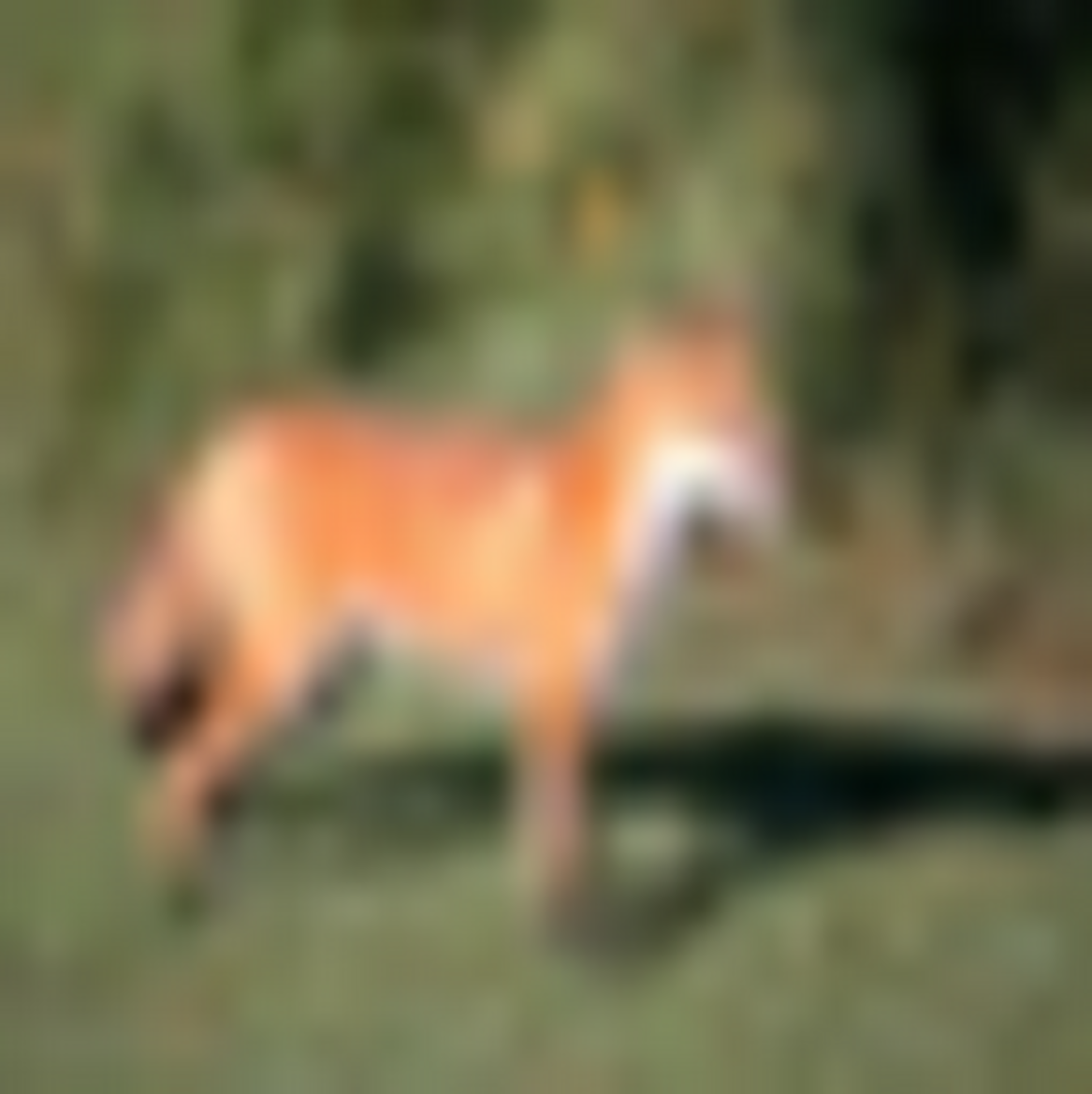}}
\hfill
\subcaptionbox{\scriptsize  Evidence 3: True label: Fox, Dist:0.664}[0.2\textwidth]{%
  \includegraphics[width=0.2\textwidth]{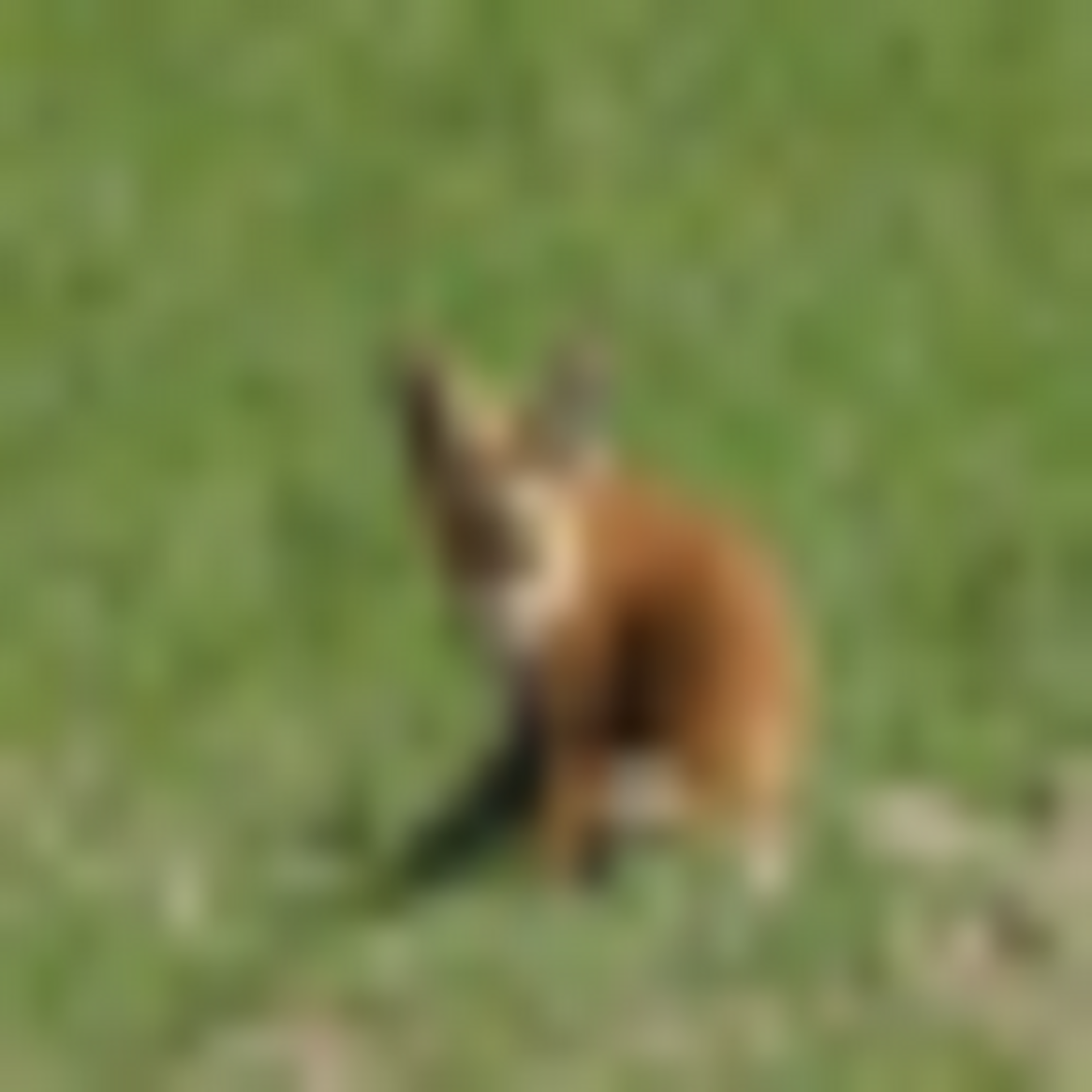}}
\\[1em]

\subcaptionbox{\scriptsize TC decision, Predicted label: Shark, True label: Shark}[0.2\textwidth]{%
  \includegraphics[width=0.2\textwidth]{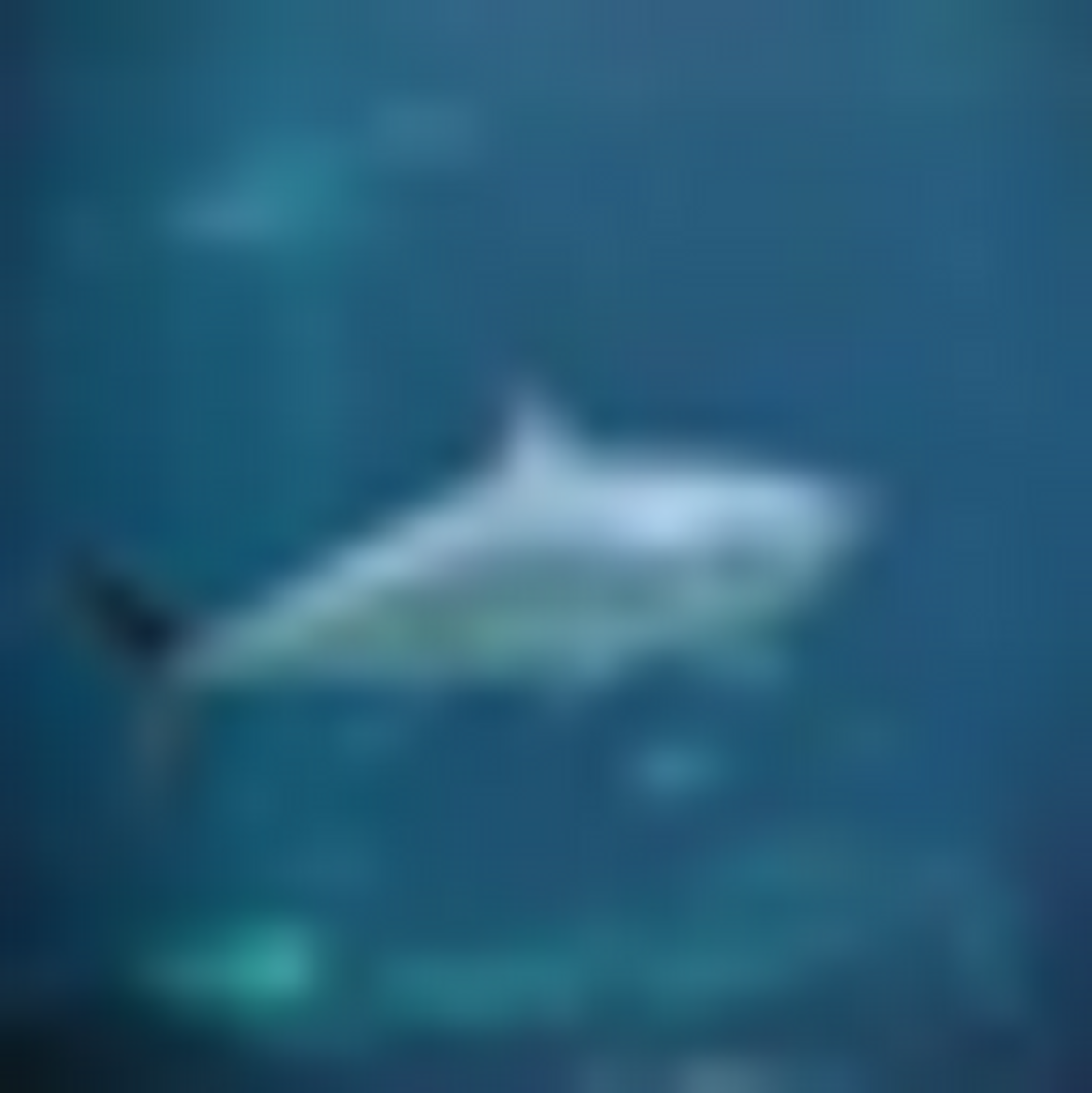}}
\hfill
\subcaptionbox{\scriptsize Evidence 1: True label: Shark, Dist: 0.0.658}[0.2\textwidth]{%
  \includegraphics[width=0.2\textwidth]{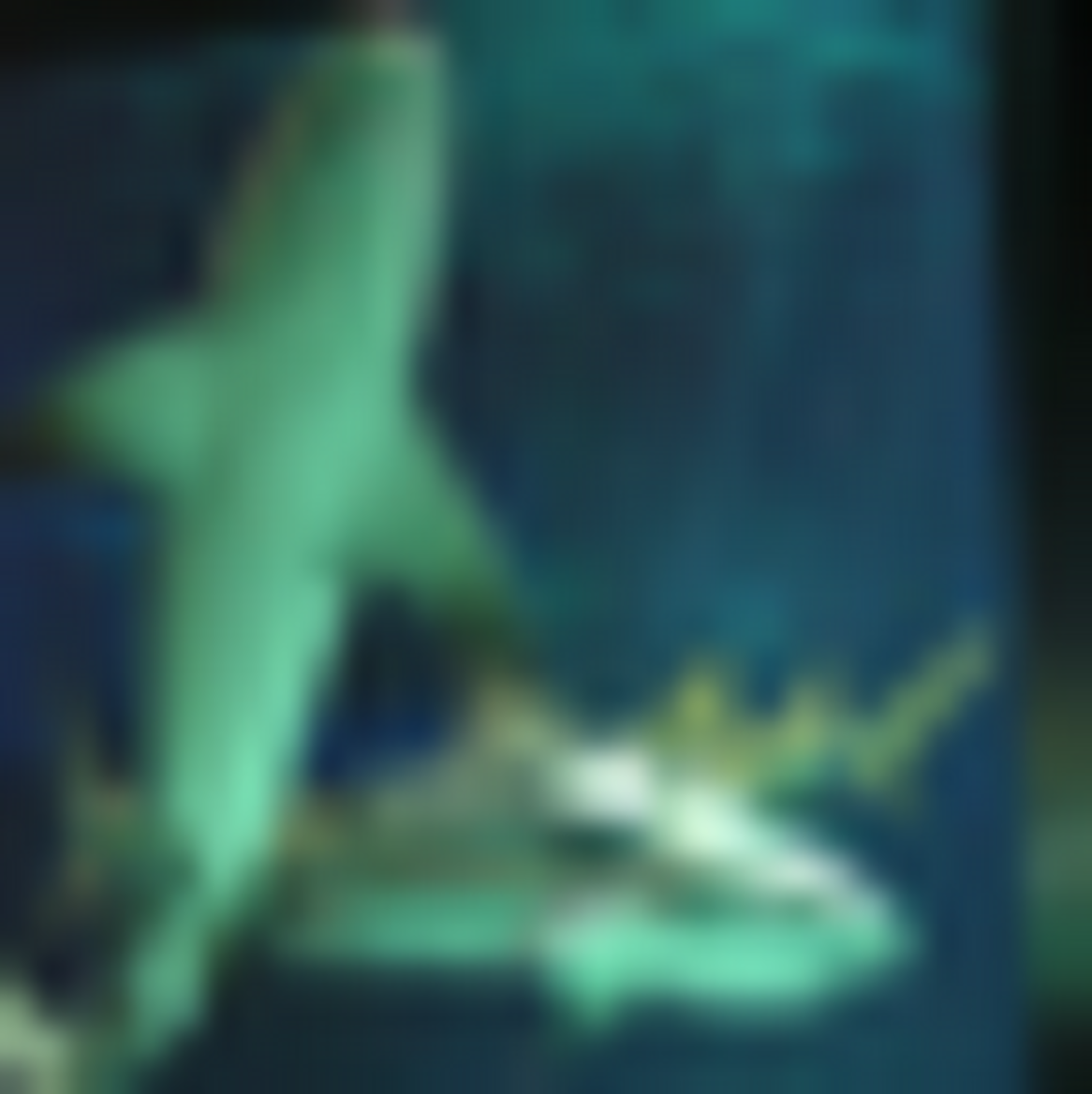}}
\hfill
\subcaptionbox{\scriptsize Evidence 1: True label: Shark, Dist: 0.0.708}[0.2\textwidth]{%
  \includegraphics[width=0.2\textwidth]{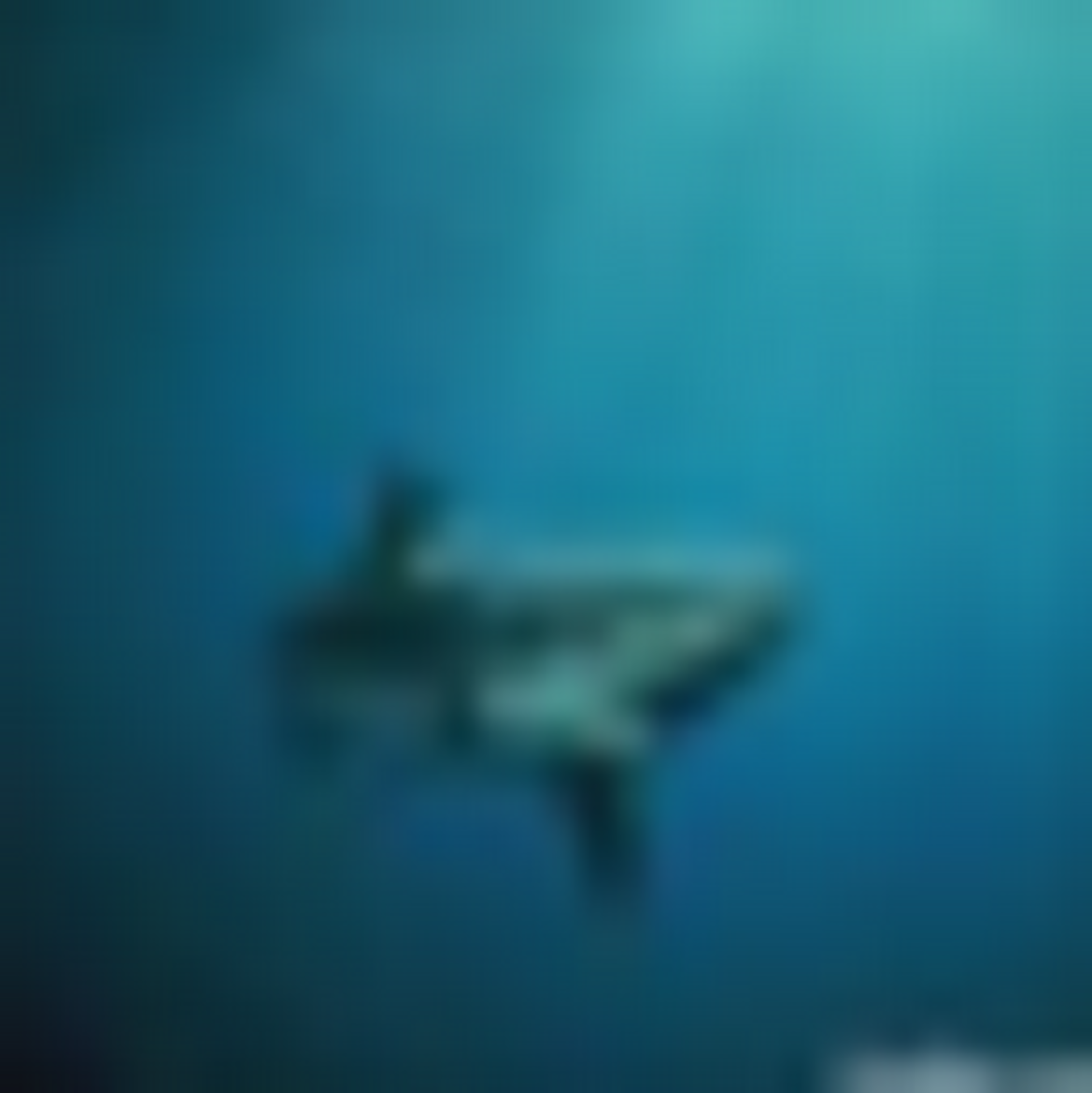}}
\hfill
\subcaptionbox{\scriptsize Evidence 1: True label: Shark, Dist: 0.771}[0.2\textwidth]{%
  \includegraphics[width=0.2\textwidth]{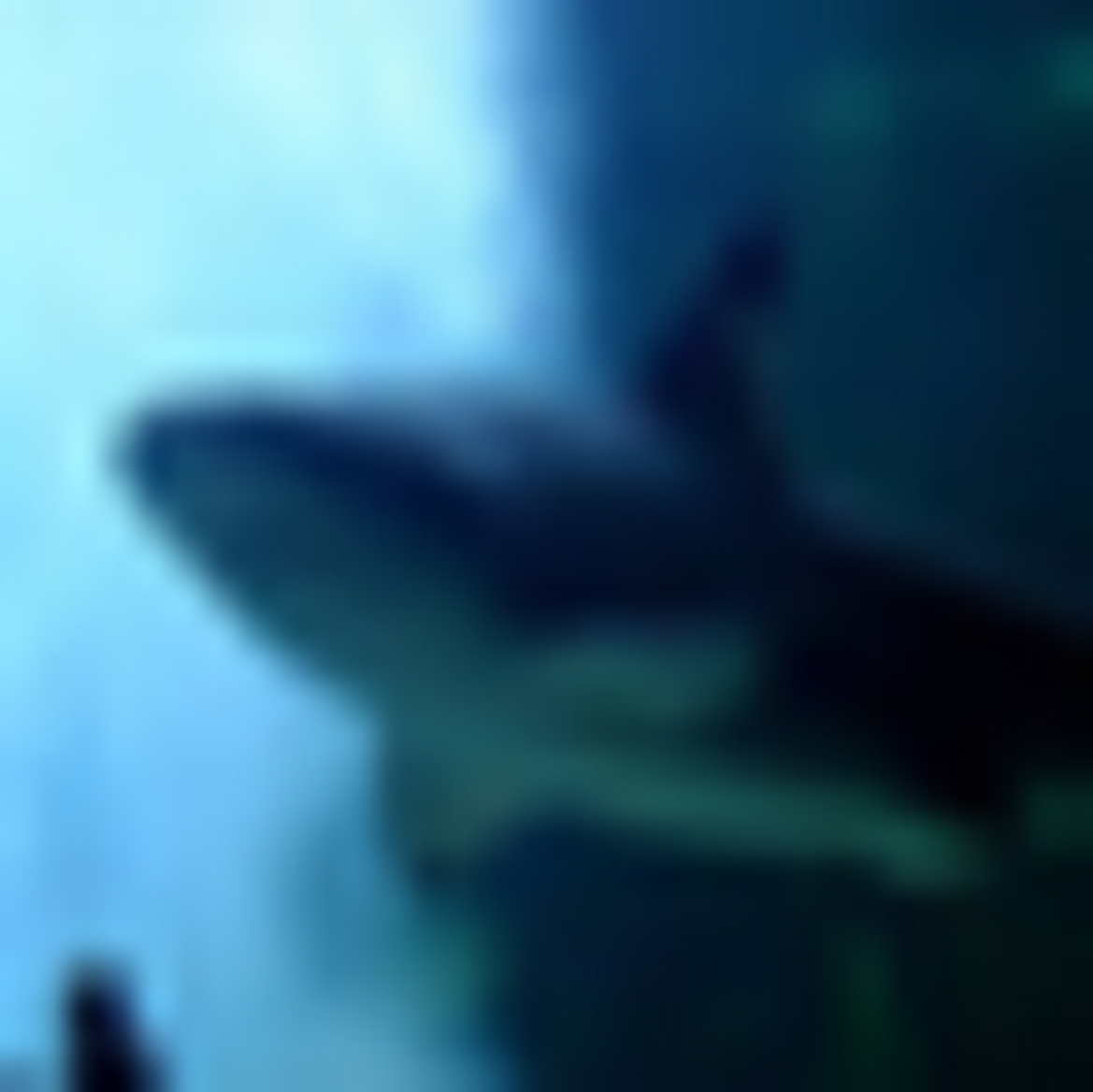}}
\\[1em]

\subcaptionbox{\scriptsize FU decision, Predicted label: Man, True label: Man}[0.2\textwidth]{%
  \includegraphics[width=0.2\textwidth]{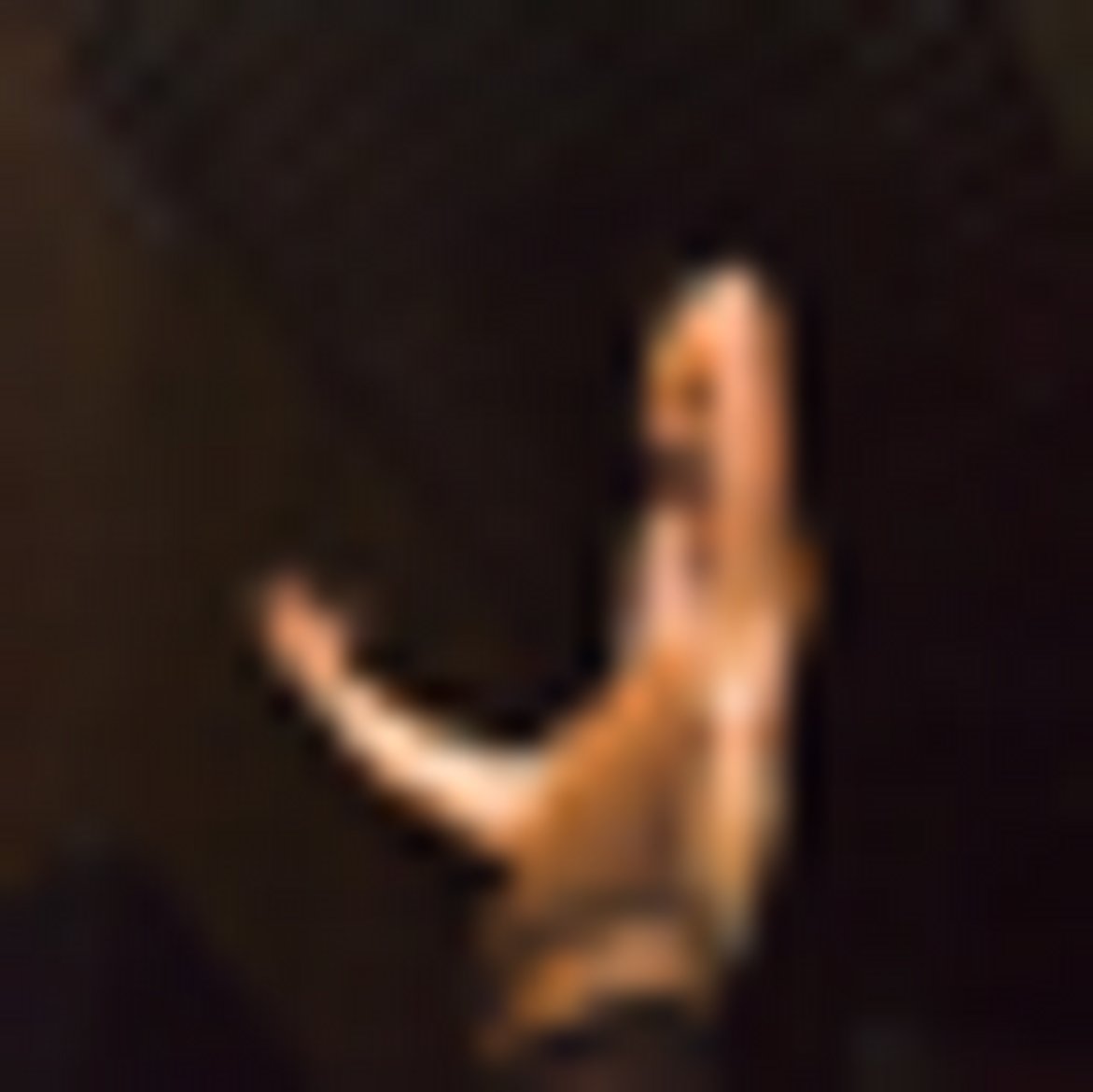}}
\hfill
\subcaptionbox{\scriptsize Evidence 1: True label:Lizard , Dist: 0.383}[0.2\textwidth]{%
  \includegraphics[width=0.2\textwidth]{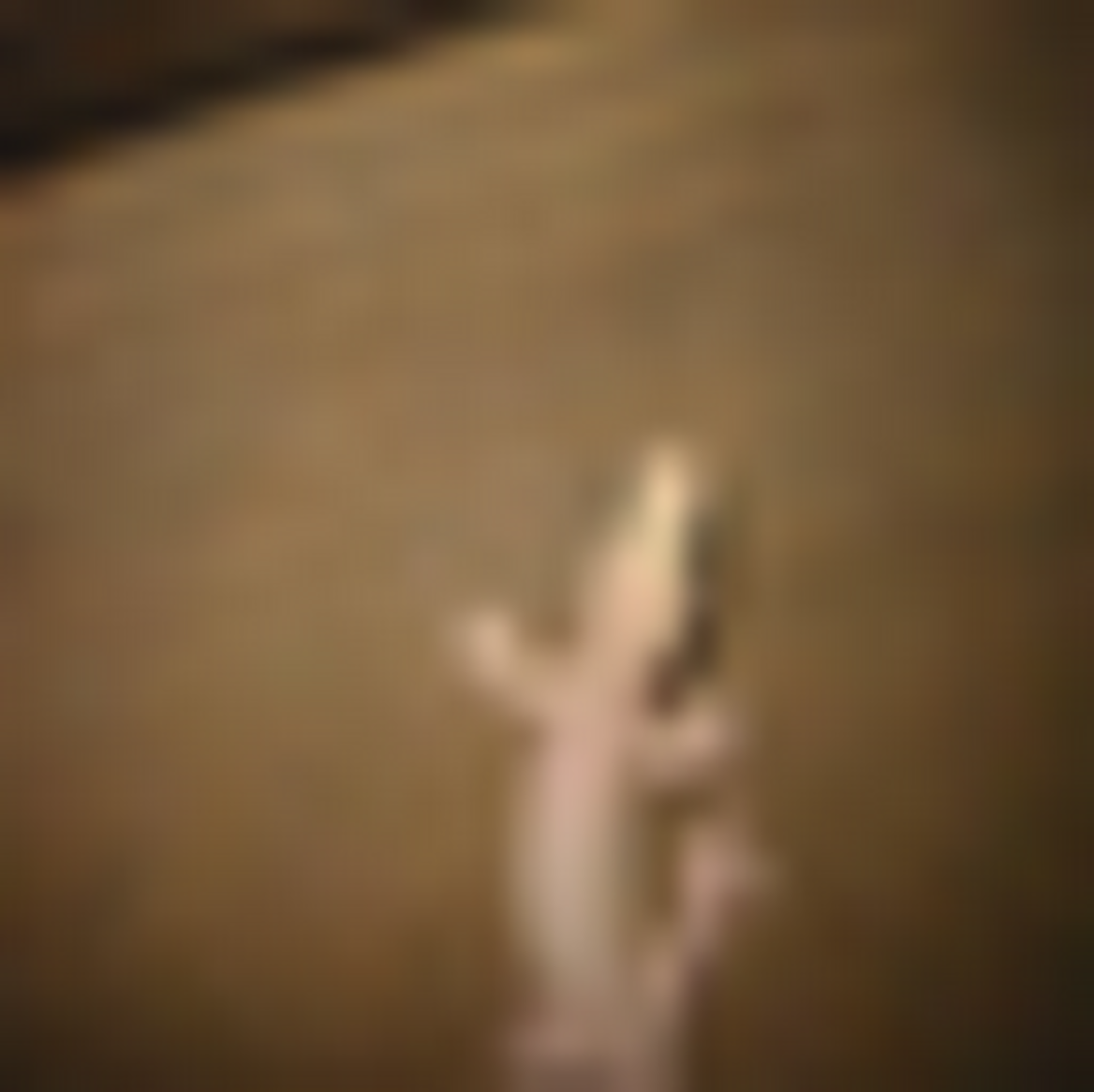}}
\hfill
\subcaptionbox{\scriptsize Evidence 1: True label: Man, Dist: 0.444}[0.2\textwidth]{%
  \includegraphics[width=0.2\textwidth]{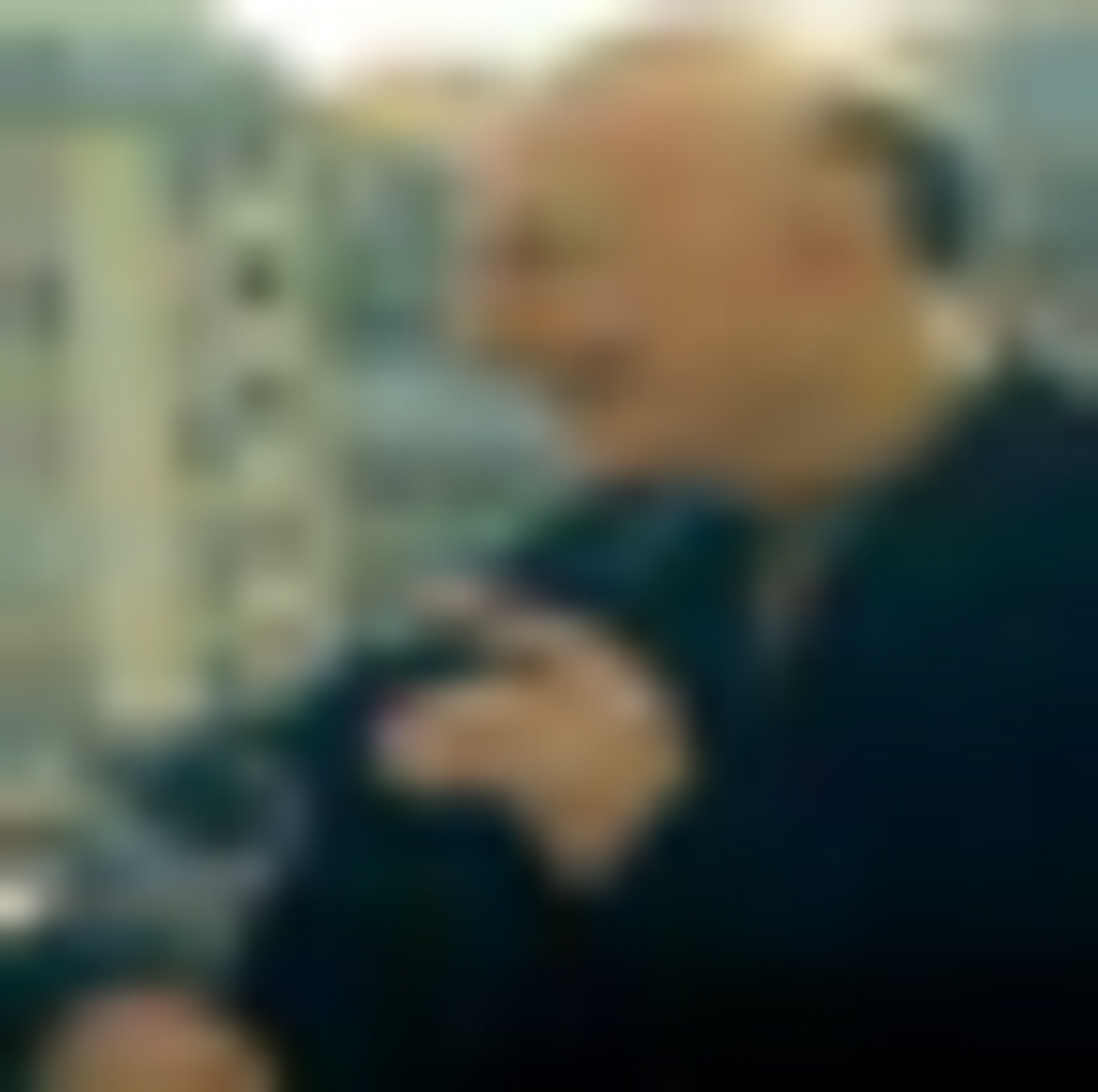}}
\hfill
\subcaptionbox{\scriptsize Evidence 1: True label: Man, Dist: 0.509}[0.2\textwidth]{%
  \includegraphics[width=0.2\textwidth]{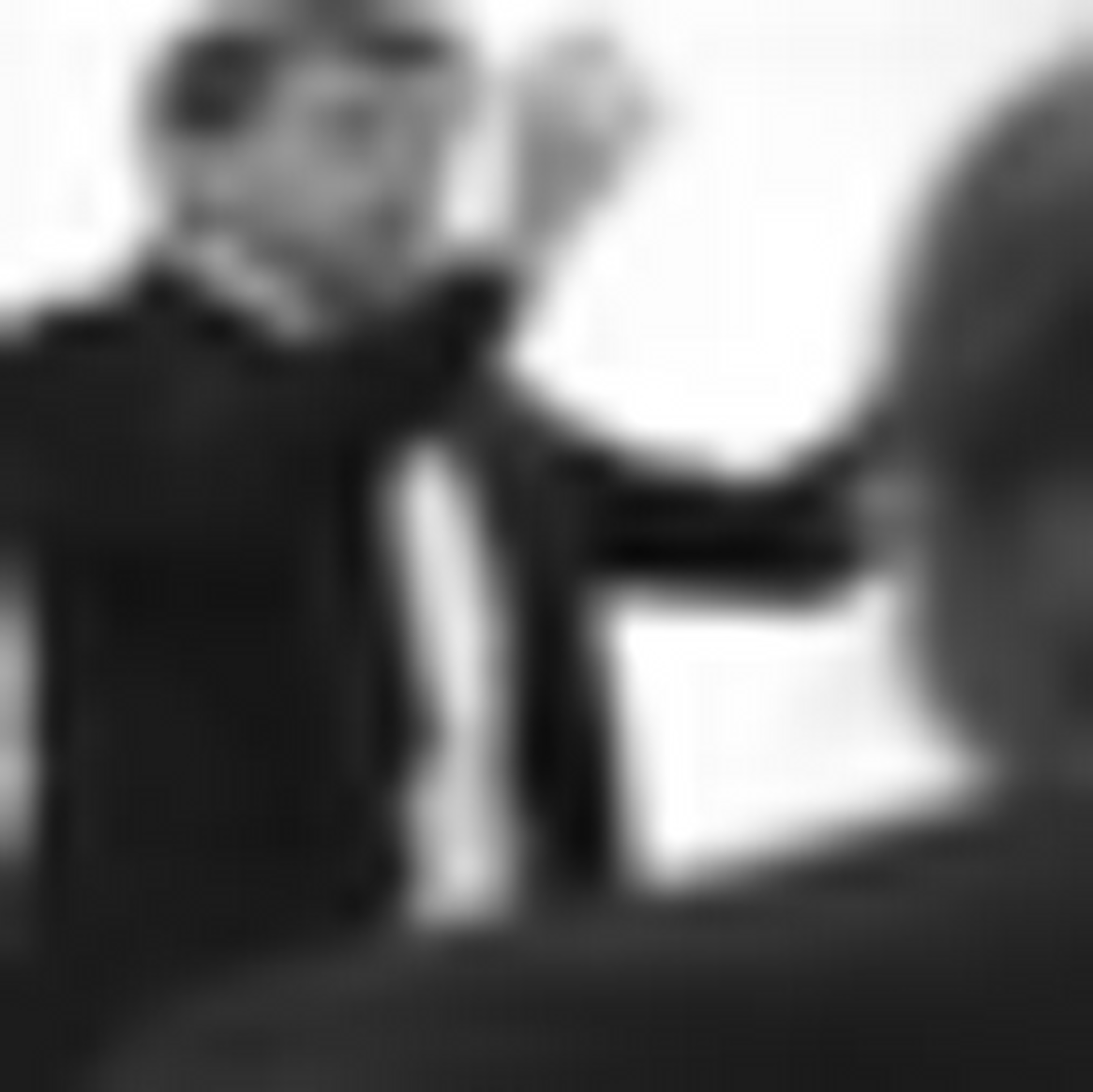}}
\\[1em]

\subcaptionbox{\scriptsize TU decision, Predicted label: Ray, True label: Lobster}[0.2\textwidth]{%
  \includegraphics[width=0.2\textwidth]{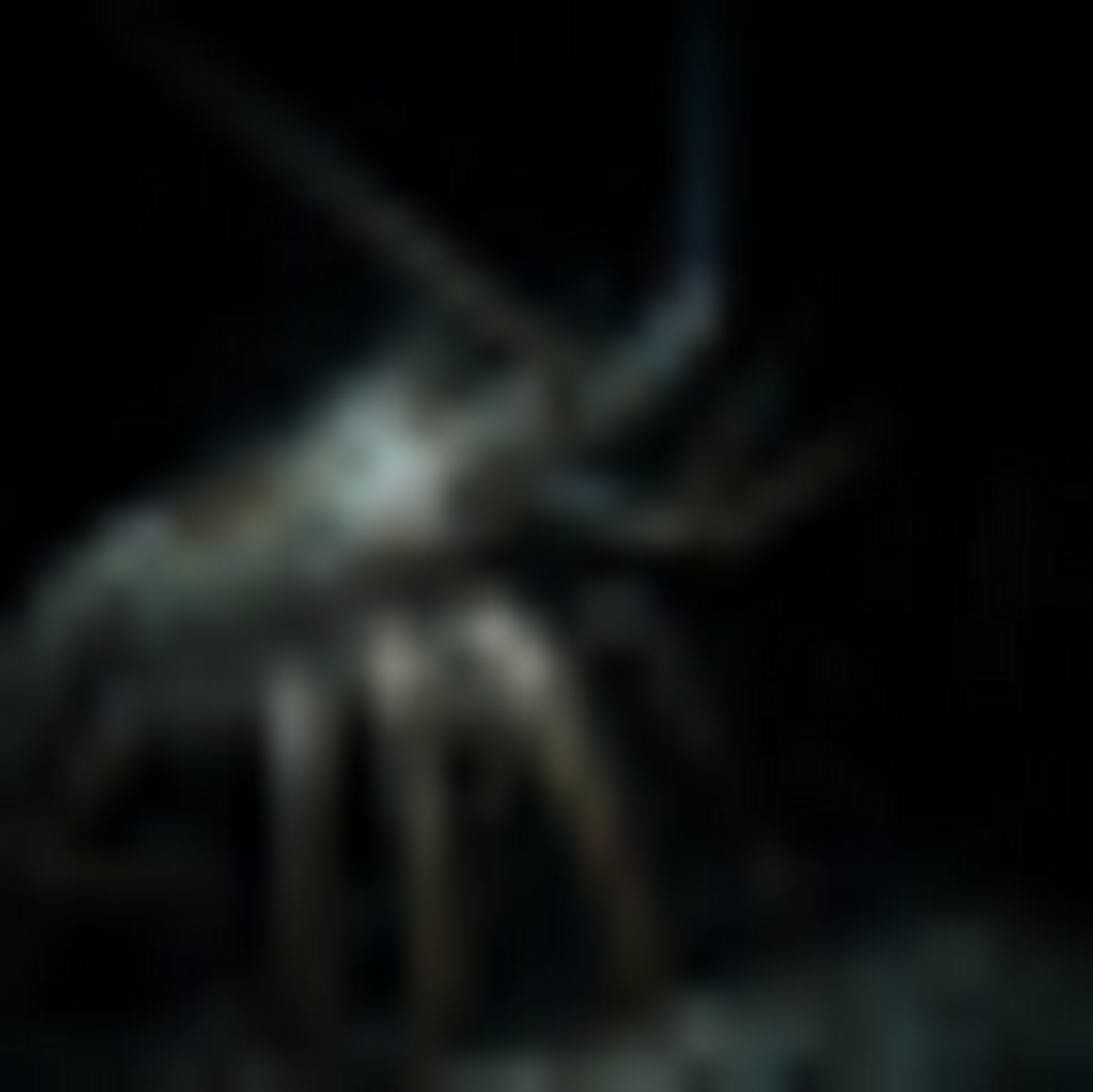}}
\hfill
\subcaptionbox{\scriptsize Evidence 1: True label: Lobster, Dist: 0.430}[0.2\textwidth]{%
  \includegraphics[width=0.2\textwidth]{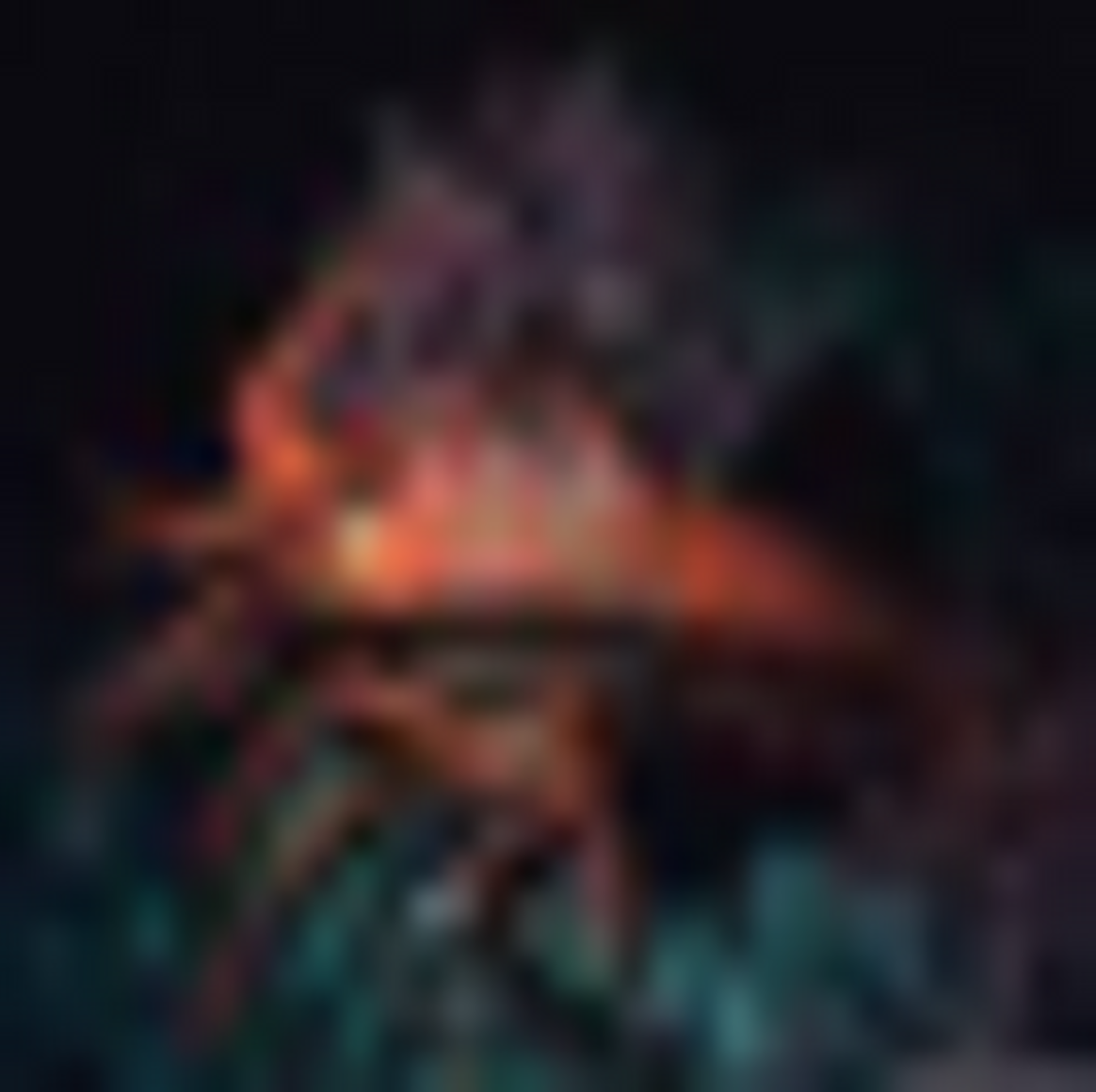}}
\hfill
\subcaptionbox{\scriptsize Evidence 1: True label: Aquarium fish, Dist: 0.442}[0.2\textwidth]{%
  \includegraphics[width=0.2\textwidth]{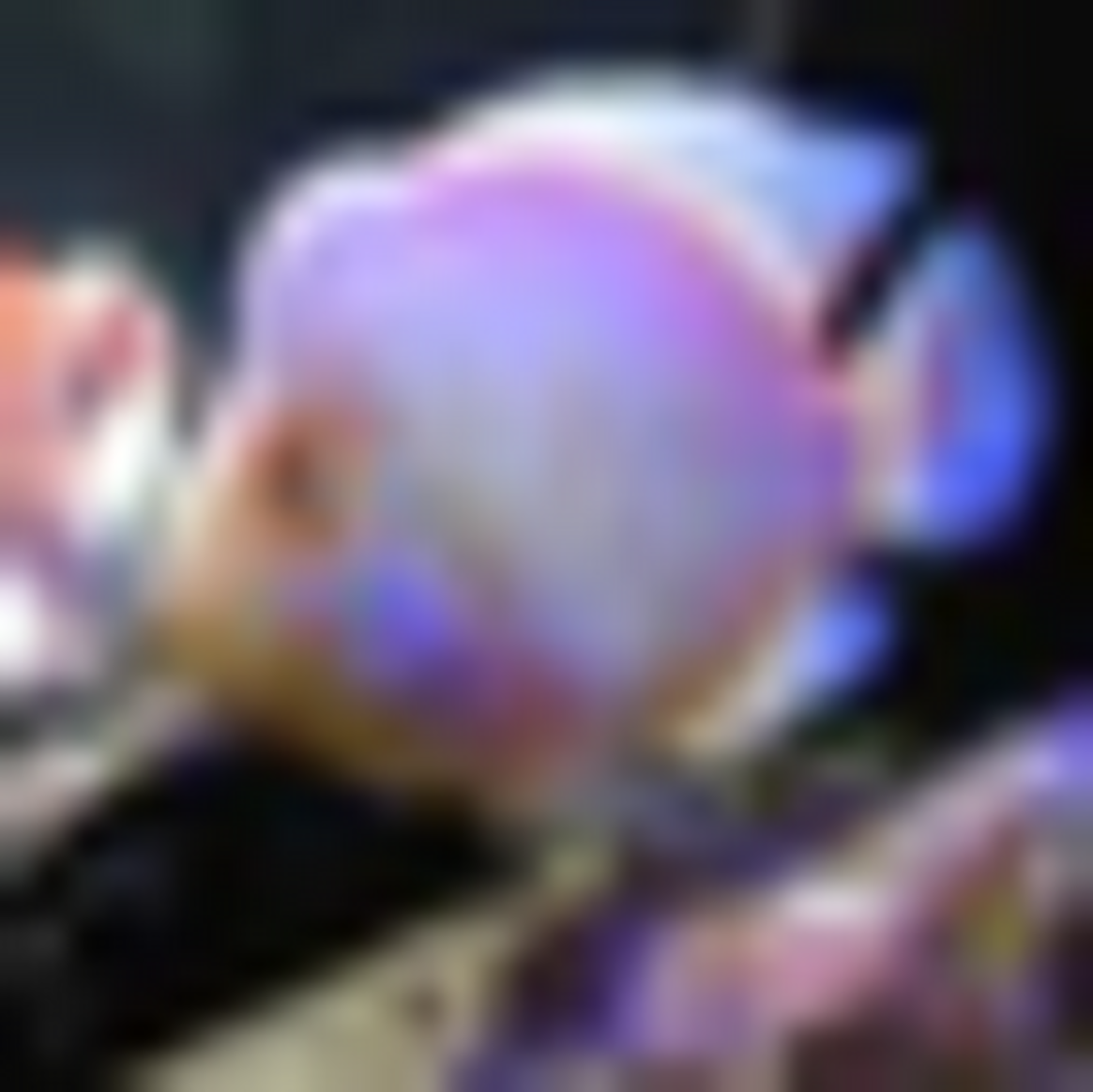}}
\hfill
\subcaptionbox{\scriptsize Evidence 1: True label: Spider , Dist: 0.488}[0.2\textwidth]{%
  \includegraphics[width=0.2\textwidth]{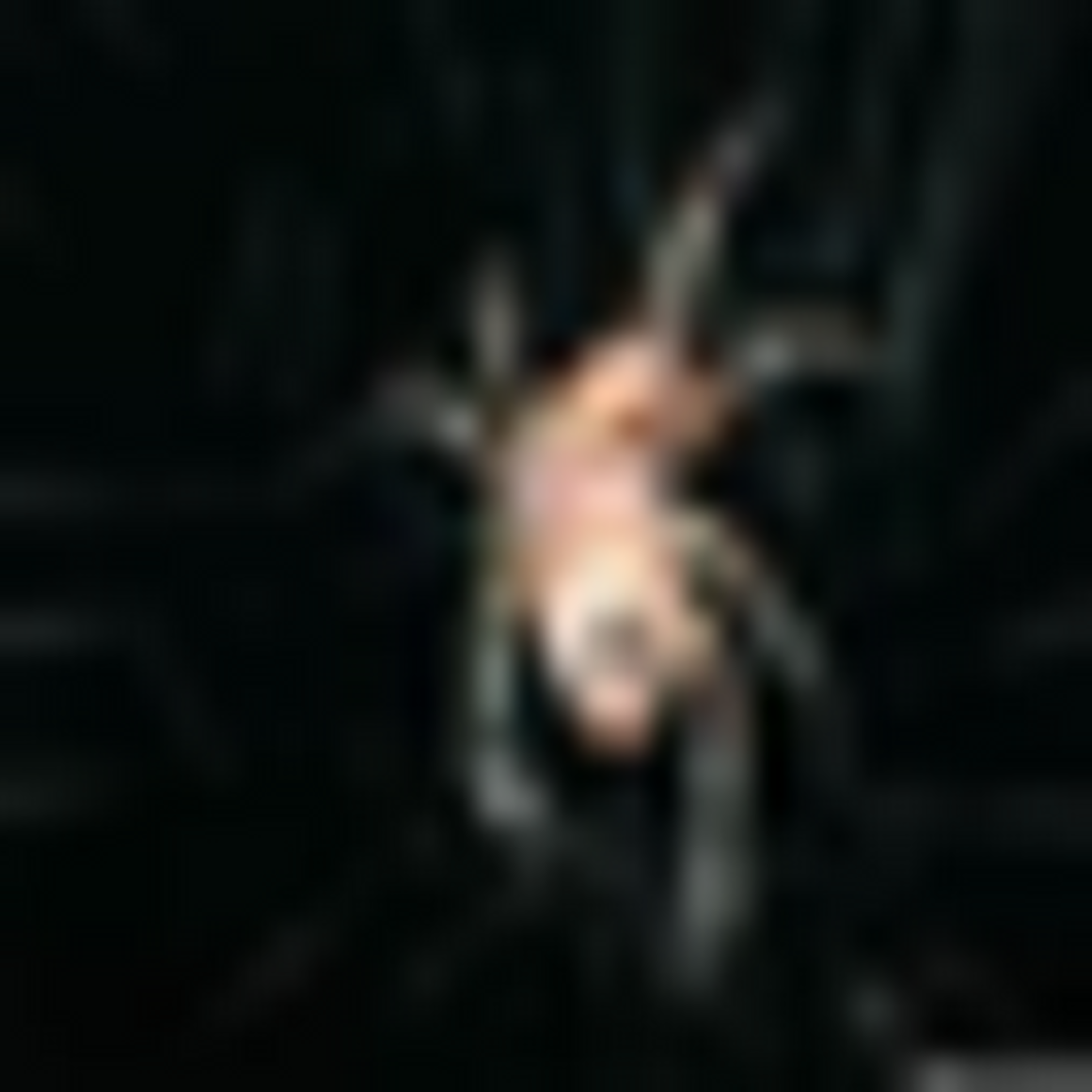}}

\caption{Examples of uncertainty-aware decision making via evidence retrieval. Rows correspond to decision categories (TC, FC, TU, FU); within each row, the left image is the test sample (true/predicted labels), followed by the top-$k=3$ nearest evidences from the historical evidence set, retrieved by proximity in the embedding space and shown with their labels and distances.}
\label{fig:sample_decisions}
\end{figure*}

These findings further reinforce the superiority of the proposed method as they demonstrate its ability to maintain higher reliability while avoiding inefficiencies associated with excessive deferrals. Beyond performance gains, the evidence-based nature of the method enables users to review the retrieved samples underpinning each decision, offering interpretability and traceability. This capability becomes especially important in safety-critical settings, where transparent decisions linked to past cases enhance trust and accountability. For instance, illustrative outcomes of the proposed evidence-retrieval scheme for CIFAR-100 with ViT backbone are shown in Figure \ref{fig:sample_decisions}. In the FC example (a–d), a test image with ground-truth lion is confidently predicted as fox; all three nearest evidences are fox with small embedding distances ($\approx 0.63–0.66$), indicating that the sample is located inside the fox neighborhood and explaining the misplaced certainty through class overlap/visual ambiguity. In the TC example (e–h), a shark image is correctly predicted, and all evidences are also shark ($\approx 0.66–0.77$), so high belief is supported by consistent neighbors. In the FU example (i–l), a man image is correctly predicted but flagged uncertain because the closest evidence belongs to a different class (lizard), while the other evidences are man; the mixed support lowers belief. In the TU example (m–p), an image with true label lobster is misclassified as ray while the evidences are heterogeneous (lobster, aquarium fish, spider), yielding low belief and an uncertain outcome. Overall, certainty is observed when evidential neighbors align with the prediction, whereas disagreement or heterogeneous neighborhoods trigger uncertainty; the retrieved samples thereby make the basis of each decision transparent and auditable.

\subsection{Ablation Study}
\begin{figure*}
\centering
\subcaptionbox{\scriptsize FC vs Belief Threshold; k=3}[0.32\textwidth]{%
  \includegraphics[width=0.32\textwidth]{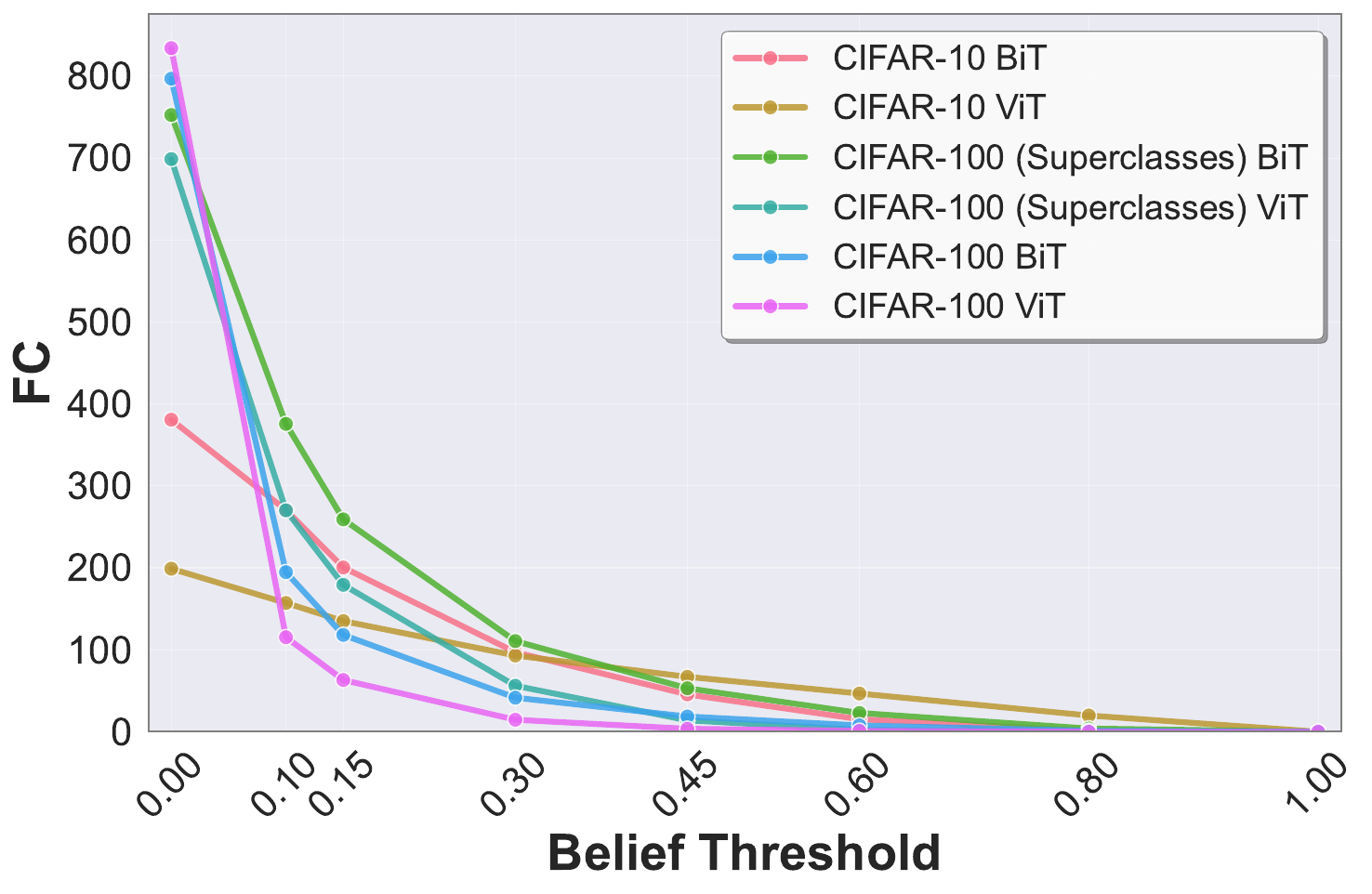}}
\hfill
\subcaptionbox{\scriptsize TC vs Belief Threshold; k=3}[0.32\textwidth]{%
  \includegraphics[width=0.32\textwidth]{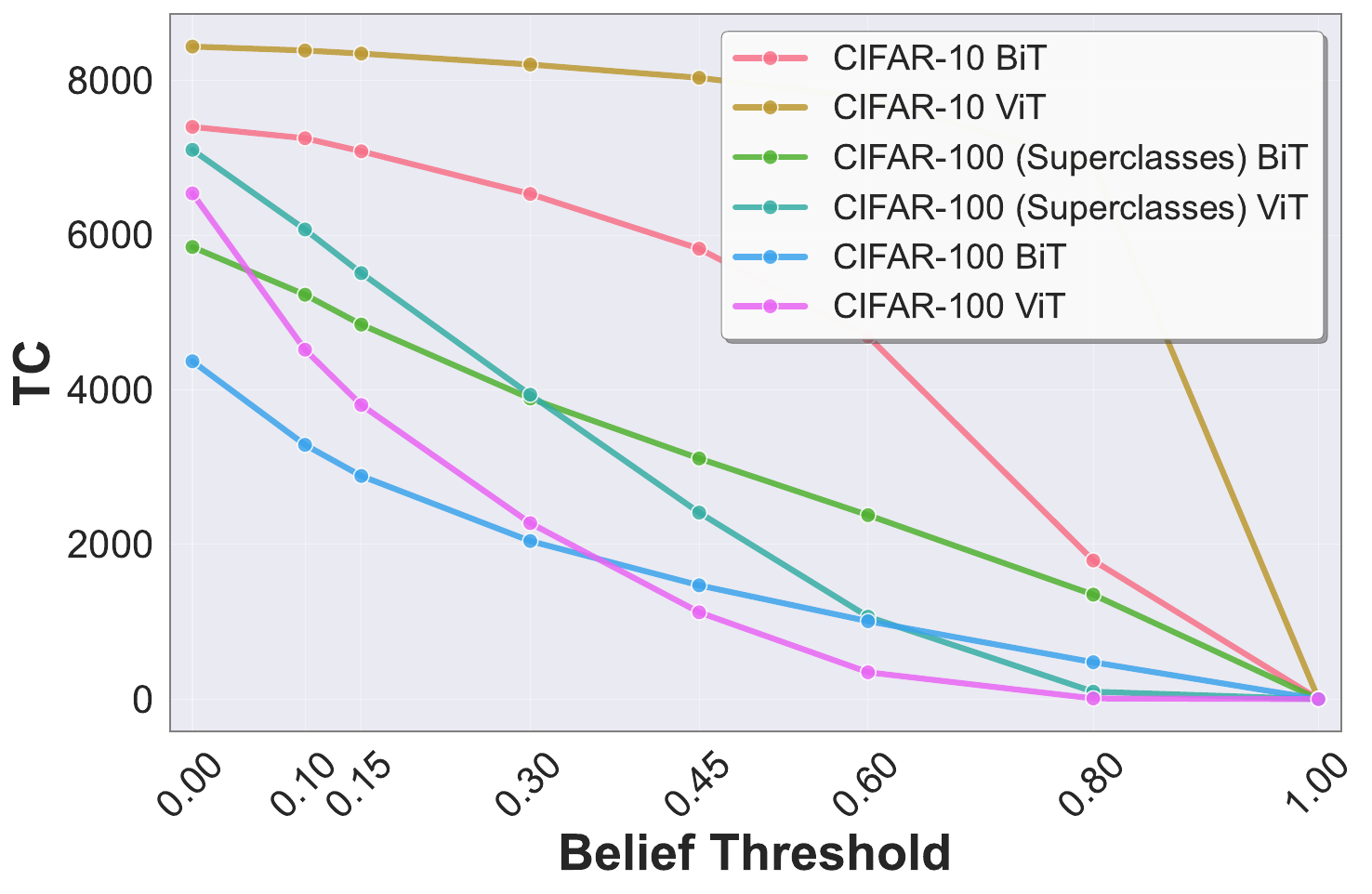}}
\hfill
\subcaptionbox{\scriptsize UG-Mean vs Belief Threshold; k=3}[0.32\textwidth]{%
  \includegraphics[width=0.32\textwidth]{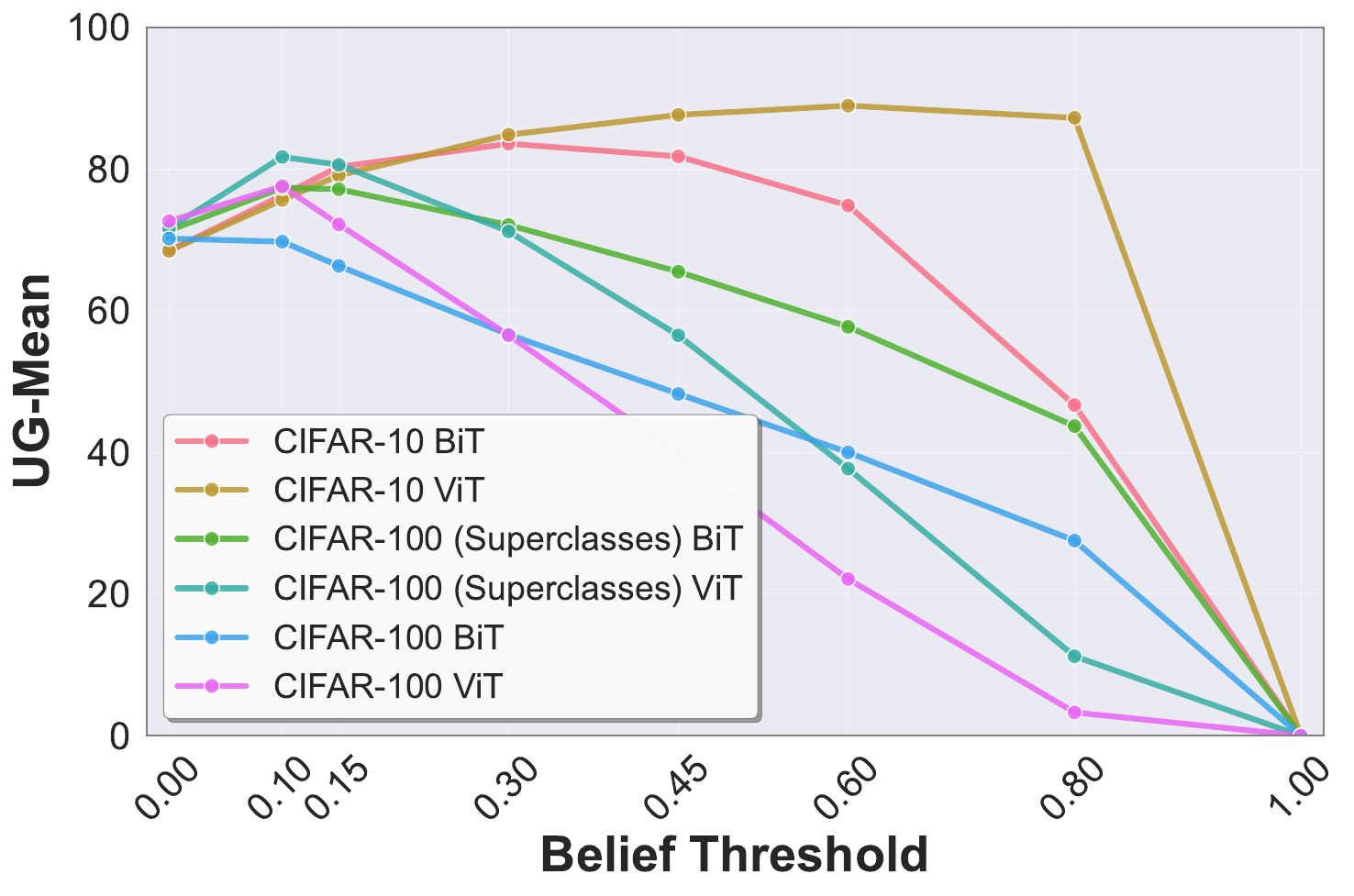}}
\\[1em]

\subcaptionbox{\scriptsize FC vs Belief Threshold; k=10}[0.32\textwidth]{%
  \includegraphics[width=0.32\textwidth]{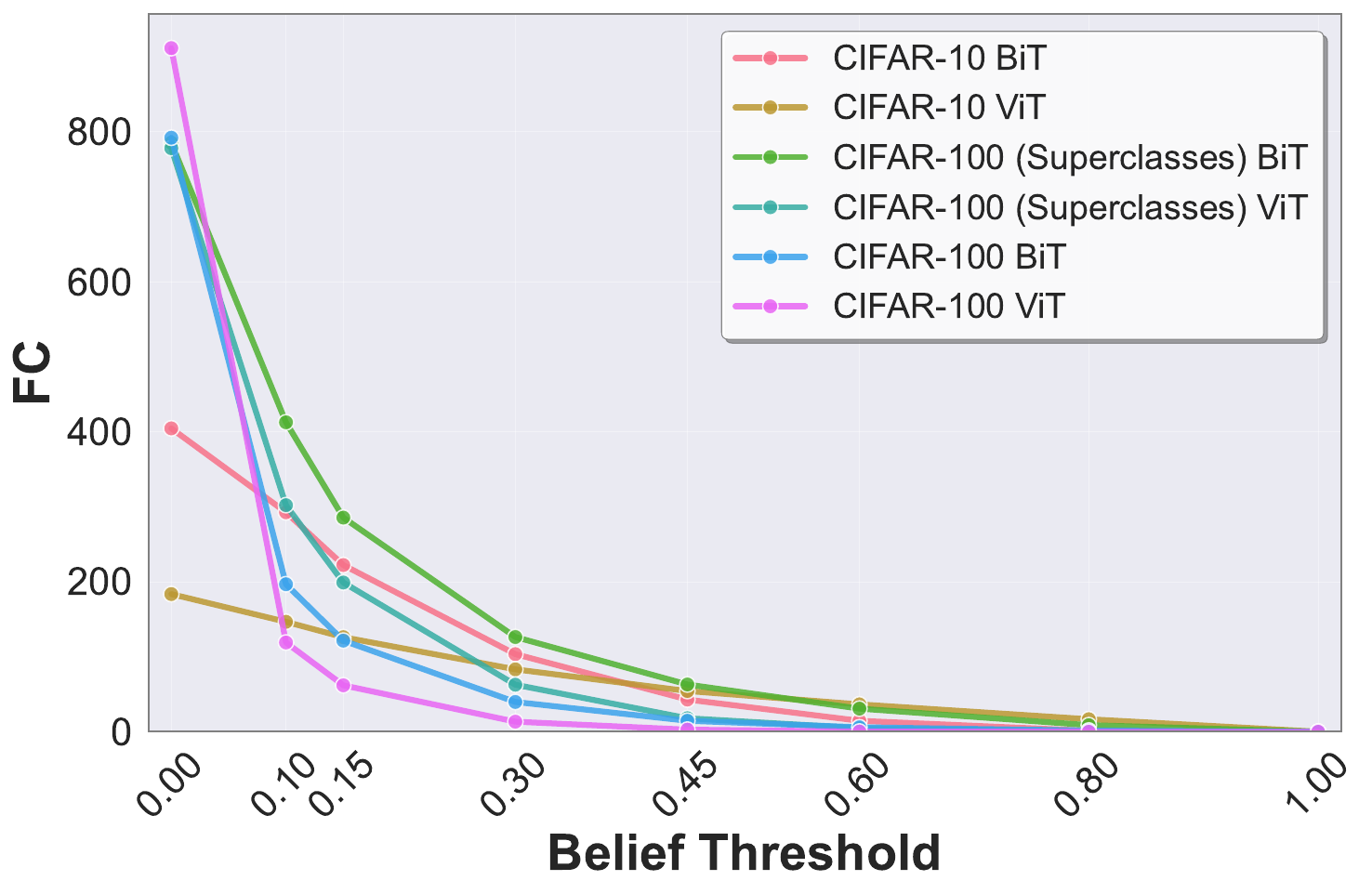}}
\hfill
\subcaptionbox{\scriptsize TC vs Belief Threshold; k=10}[0.32\textwidth]{%
  \includegraphics[width=0.32\textwidth]{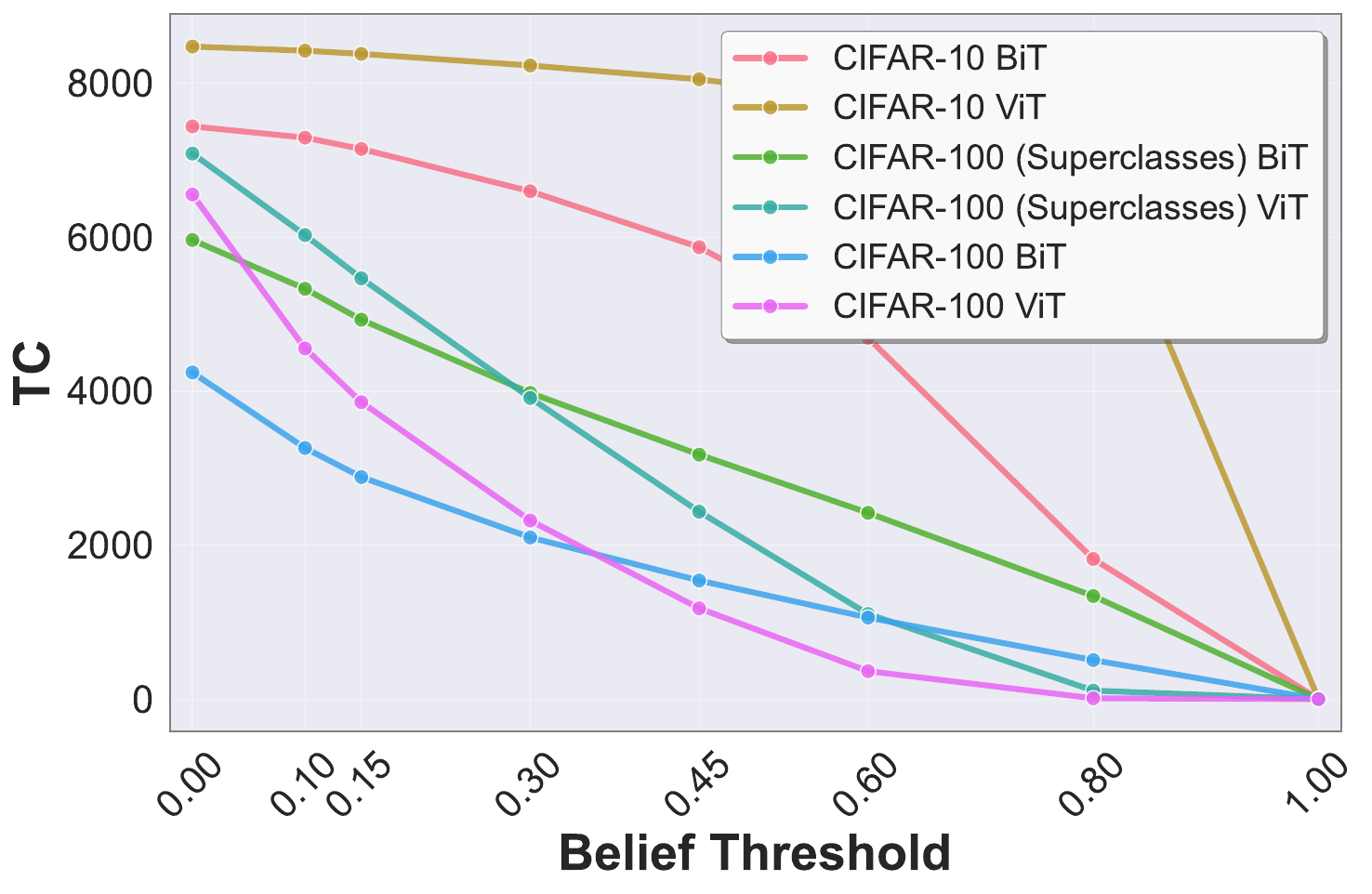}}
\hfill
\subcaptionbox{\scriptsize UG-Mean vs Belief Threshold; k=10}[0.32\textwidth]{%
  \includegraphics[width=0.32\textwidth]{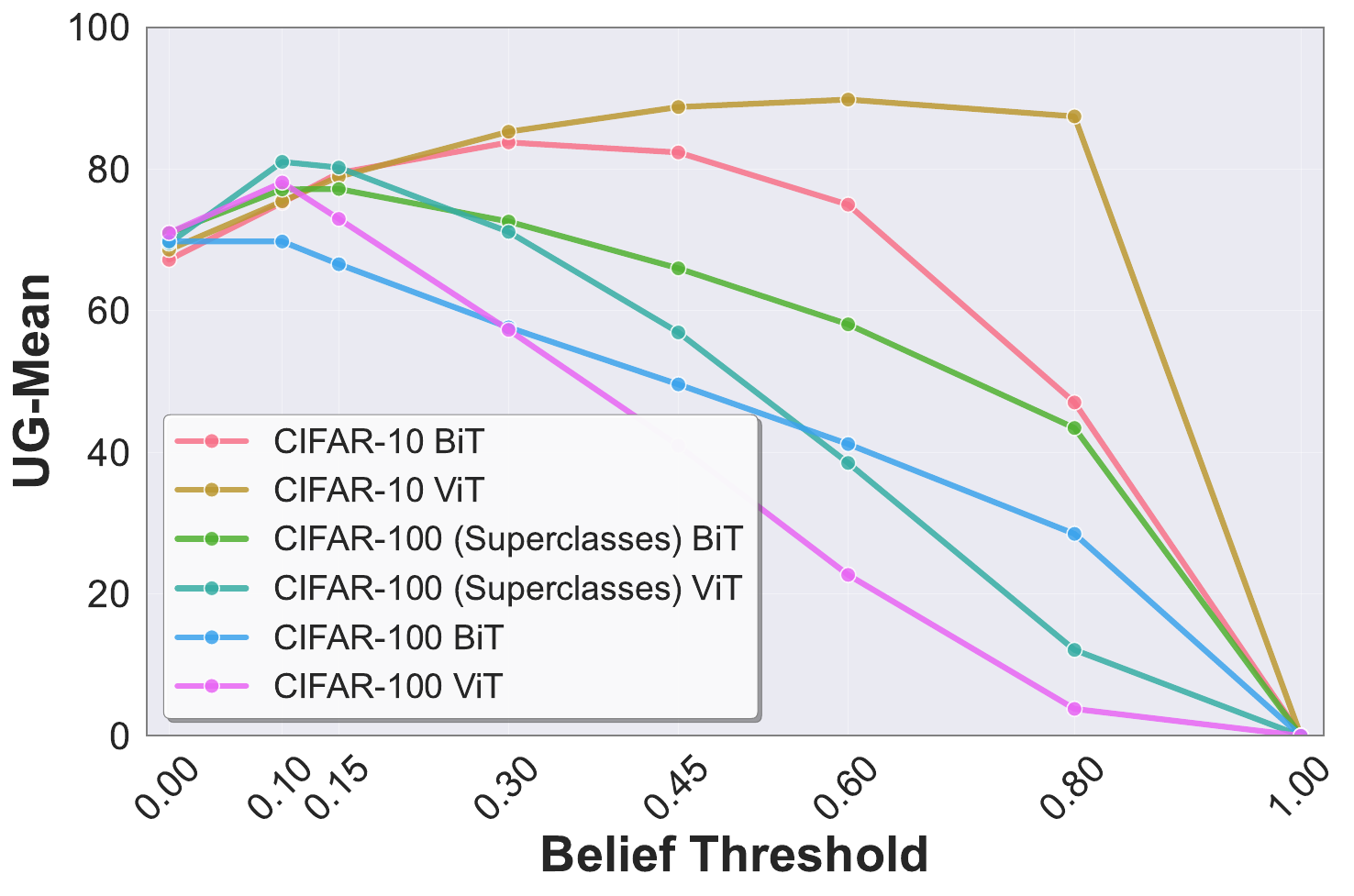}}
\\[1em]

\subcaptionbox{\scriptsize FC vs Belief Threshold; k=30}[0.32\textwidth]{%
  \includegraphics[width=0.32\textwidth]{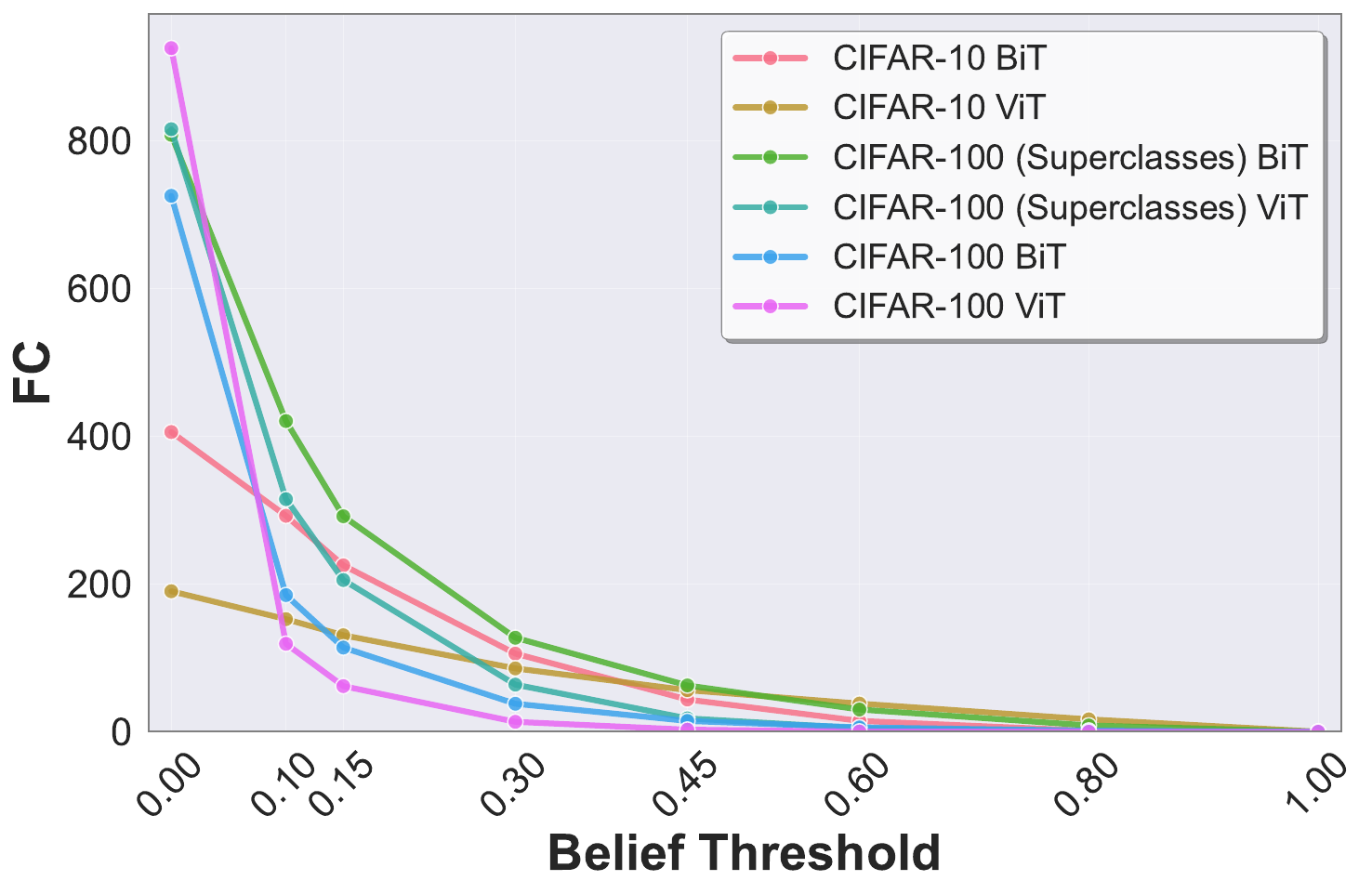}}
\hfill
\subcaptionbox{\scriptsize TC vs Belief Threshold; k=30}[0.32\textwidth]{%
  \includegraphics[width=0.32\textwidth]{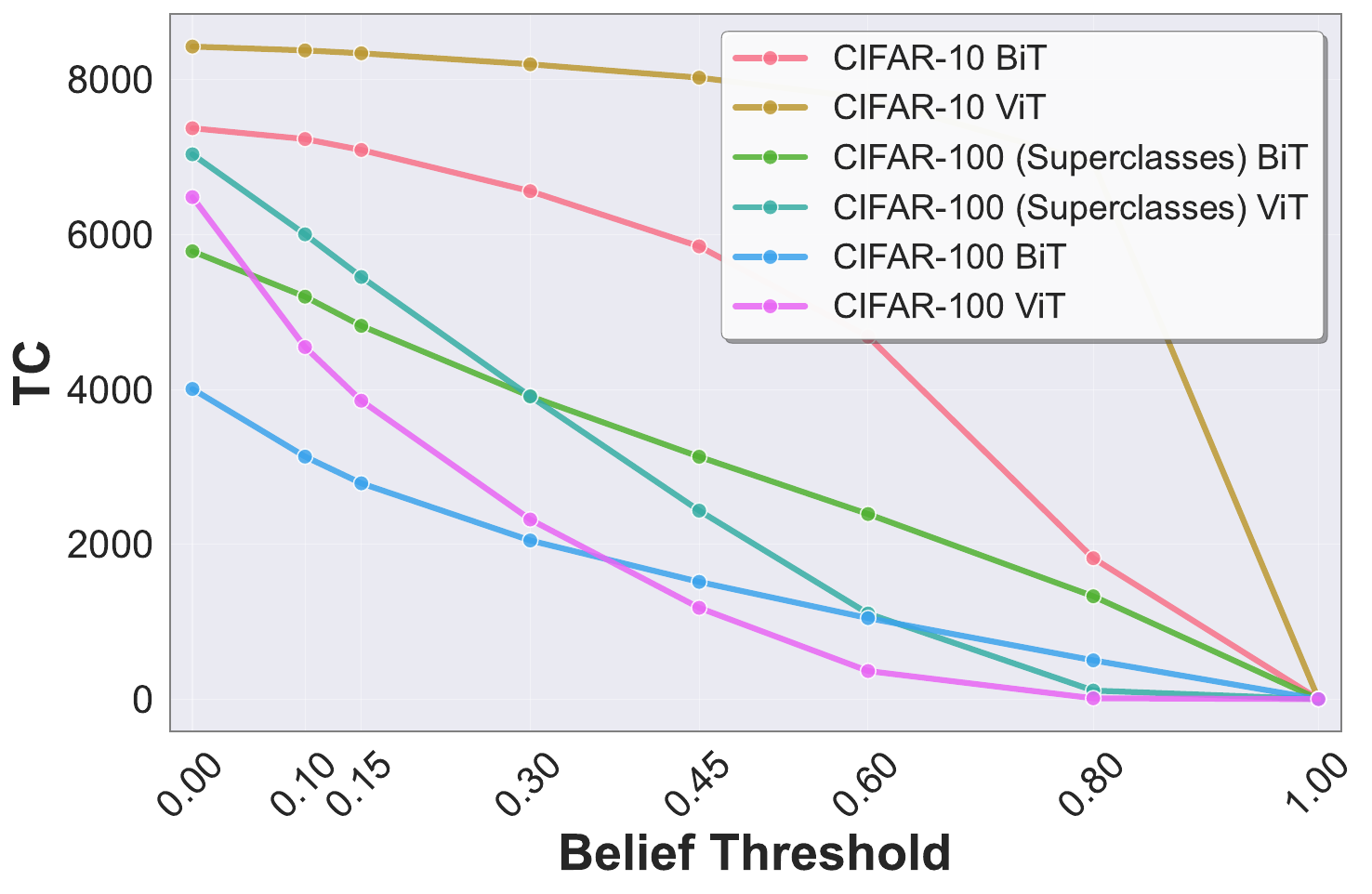}}
\hfill
\subcaptionbox{\scriptsize UG-Mean vs Belief Threshold; k=30}[0.32\textwidth]{%
  \includegraphics[width=0.32\textwidth]{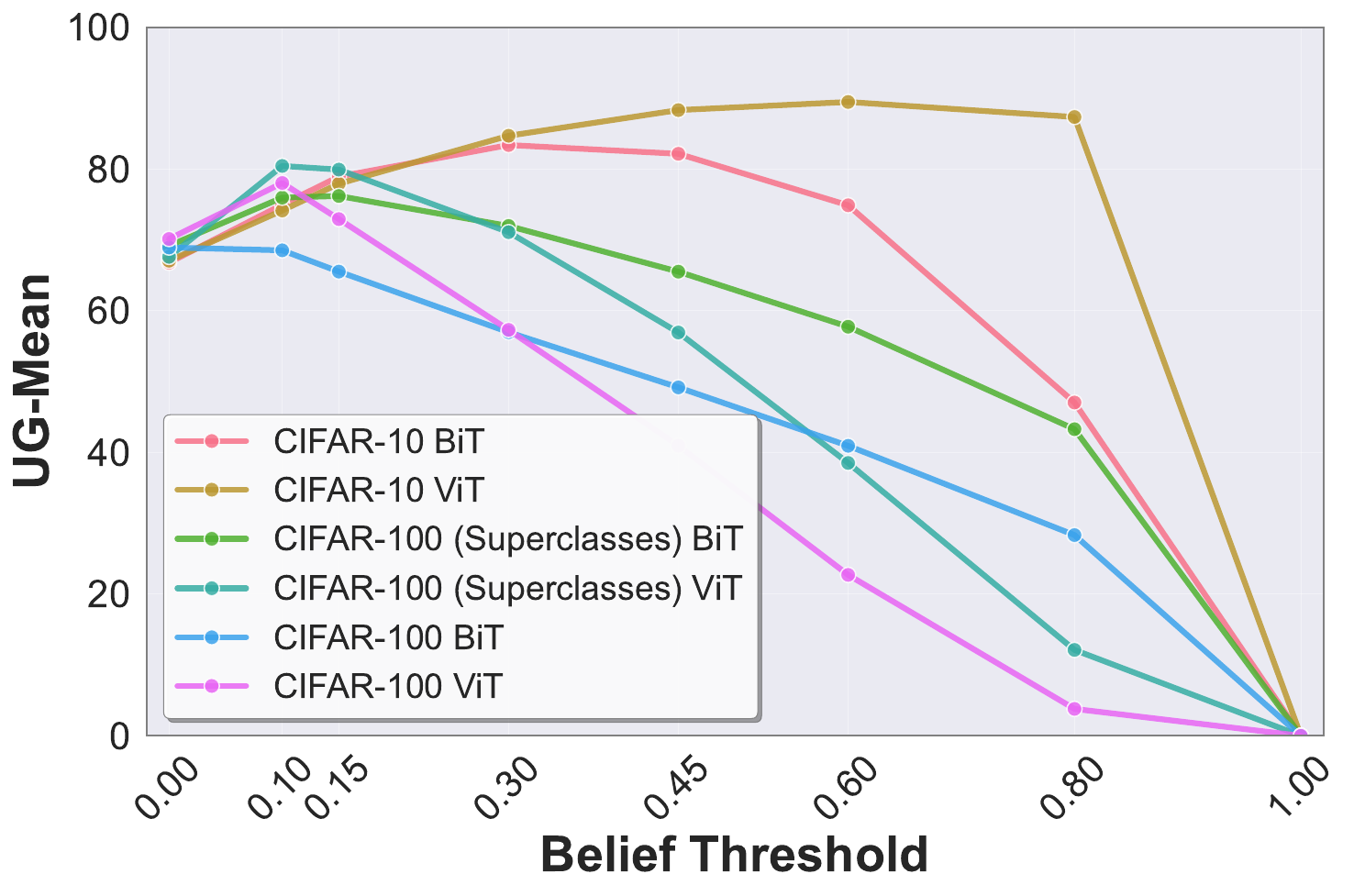}}
\\[1em]

\subcaptionbox{\scriptsize FC vs Belief Threshold; k=50}[0.32\textwidth]{%
  \includegraphics[width=0.32\textwidth]{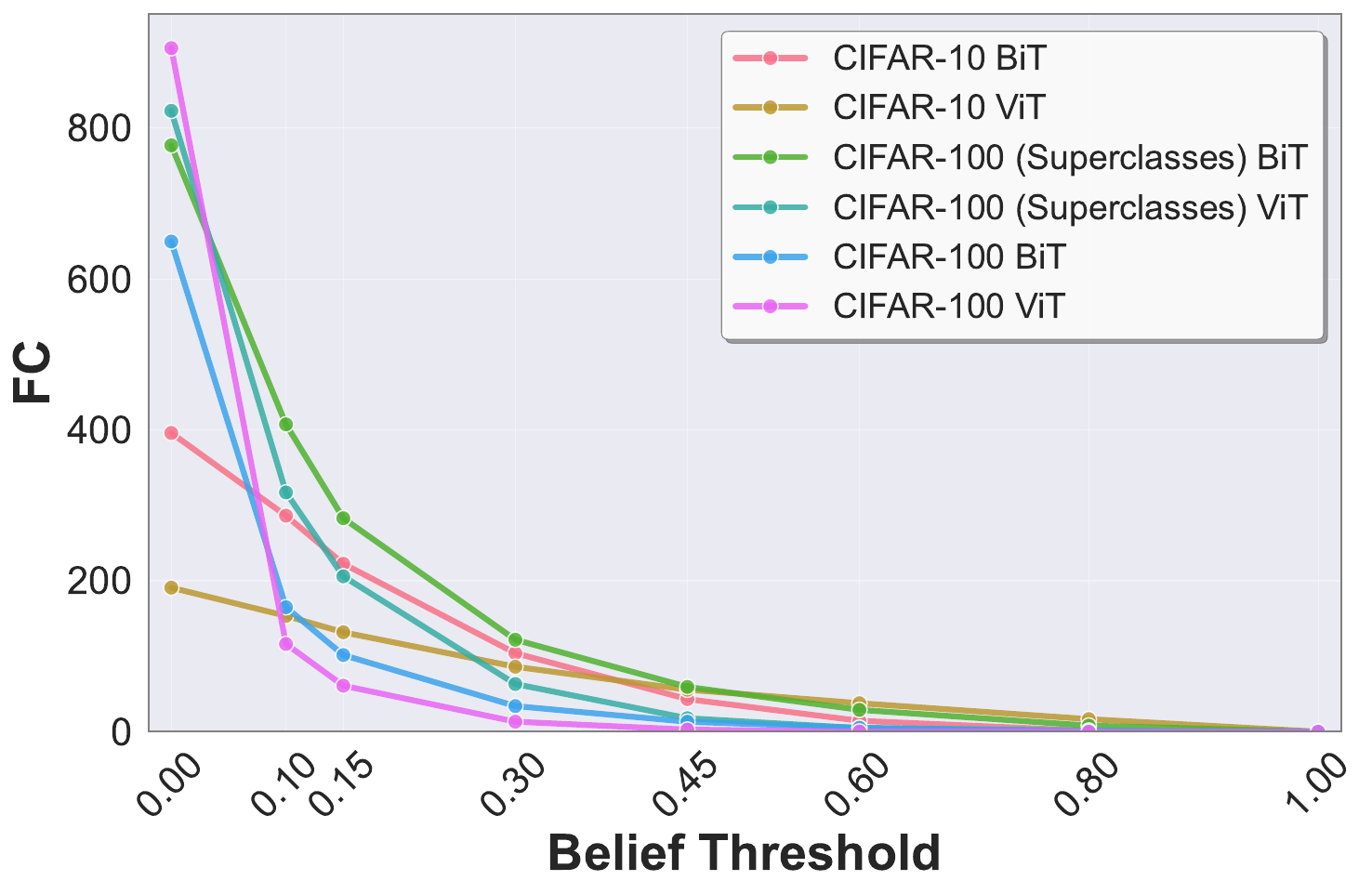}}
\hfill
\subcaptionbox{\scriptsize TC vs Belief Threshold; k=50}[0.32\textwidth]{%
  \includegraphics[width=0.32\textwidth]{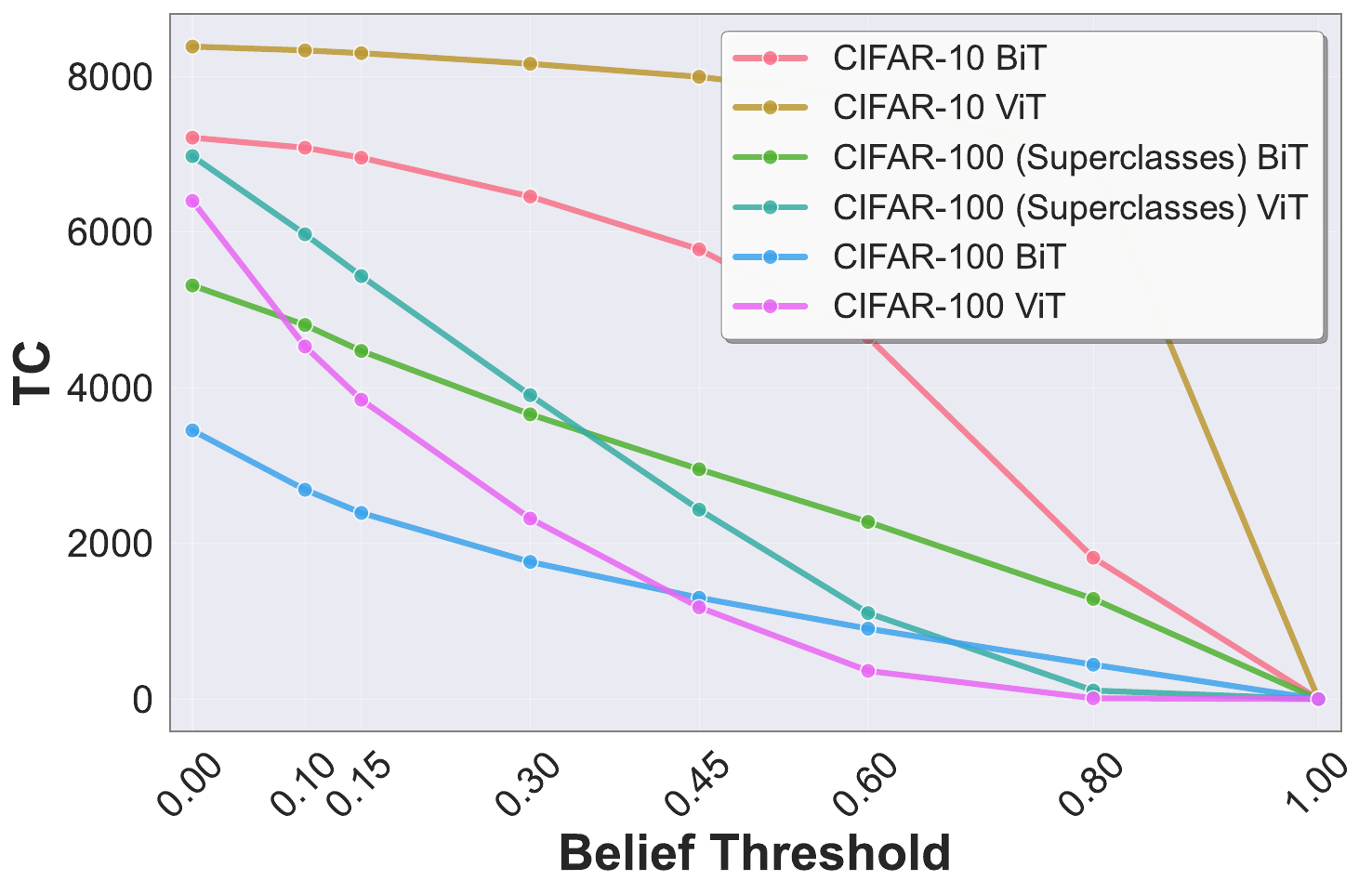}}
\hfill
\subcaptionbox{\scriptsize UG-Mean vs Belief Threshold; k=50}[0.32\textwidth]{%
  \includegraphics[width=0.32\textwidth]{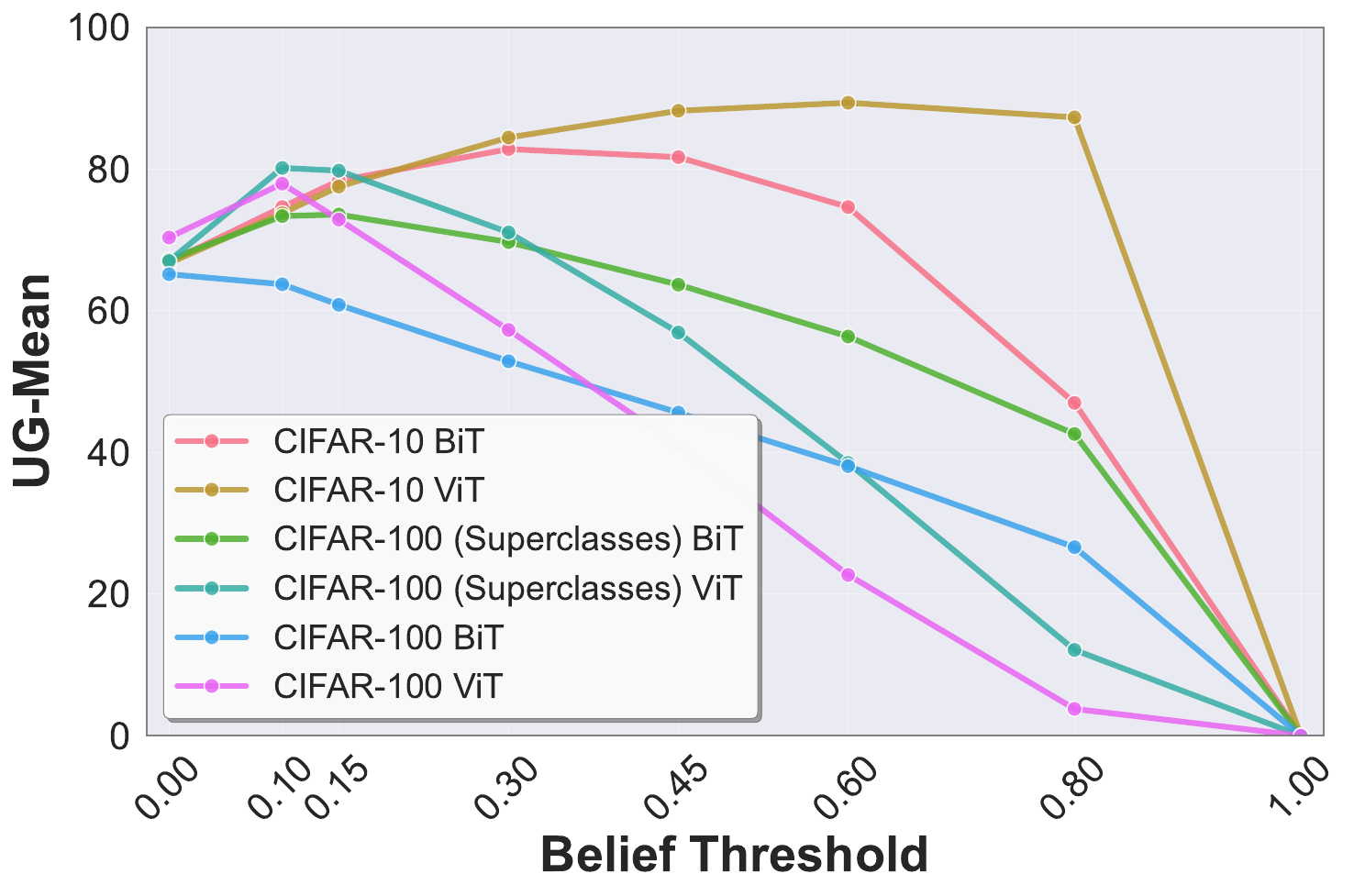}}

\caption{Parameter sensitivity analysis showing the effect of belief threshold variation on uncertainty-aware performance metrics (FC, TC, UG-Mean) for different k values and dataset-backbone combinations. Results averaged across 10 runs per configuration.}
\label{fig:sensitivity_analysis}
\end{figure*}

This section presents the parameter ablation analysis of the proposed method across different belief thresholds and k-nearest evidences configurations. The Figure \ref{fig:sensitivity_analysis} demonstrates the effect of belief threshold variation (x-axis) on three key uncertainty-aware performance metrics: FC, TC, and UG-Mean, for k values of 3, 10, 30, and 50. Each line represents a different dataset-backbone combination (CIFAR-10 BiT/ViT, CIFAR-100 Superclasses BiT/ViT, CIFAR-100 Fine-grained BiT/ViT). Results are averaged across 10 independent runs per configuration.

The ablation analysis reveals a systematic relationship between belief threshold adjustment and method behavior that follows predictable patterns. At the extremes of threshold selection ($\tau \to 0$), all predictions receive certain classifications, resulting in maximized TC but also elevated FC values that compromise safety. Conversely, as thresholds approach unity ($\tau\to 1$), the framework becomes increasingly conservative, driving both TC and FC toward zero as samples migrate to uncertain categories.

Since the proposed method does not change the predicted class—only its
certainty tag (certain vs.\ uncertain)—the total number of correct predictions and errors is fixed for a given dataset/backbone. Let \(N\) be the test size, \(T\) the number correct, and \(F=N-T\) the number incorrect. For any choice of evidence size \(k\) and belief threshold \(\tau\), 

\begin{align}
\underbrace{TC(k,\tau)}_{\text{correct \& certain}}
+\underbrace{FU(k,\tau)}_{\text{correct \& uncertain}}
&= T,\\
\underbrace{FC(k,\tau)}_{\text{wrong \& certain}}
+\underbrace{TU(k,\tau)}_{\text{wrong \& uncertain}}
&= F.
\end{align}
Hence

\begin{equation}
FU(k,\tau)=T- TC(k,\tau), \qquad
TU(k,\tau)=F- FC(k,\tau).
\end{equation}

Consequently, any decrease in \(TC\) (as \(\tau\) increases) is matched
one-for-one by an increase in \(FU\); and any decrease in \(FC\) is matched one-for-one by an increase in \(TU\). \(TU\) and \(FU\) therefore carry no independent information once \(TC\) and \(FC\) are shown, which is why they were omitted from the plots.

The UG-Mean curves exhibit characteristic inverted-U shapes that reflect the optimization of competing objectives. At low thresholds, the geometric mean suffers from elevated FC values that penalize the safety component despite high TC counts. As thresholds increase, FC suppression initially improves UG-Mean performance by reducing the false positive rate in the uncertainty detection formulation. However, continued threshold elevation eventually diminishes TC values sufficiently to offset these safety gains, creating the observed performance peaks at intermediate threshold values. This pattern emerges consistently across all experimental configurations, indicating that the optimal operating region represents a fundamental trade-off between operational efficiency and system safety rather than dataset-specific phenomena.

Considering the evidence size impact, k, the evidence aggregation analysis reveals diminishing returns from increased evidences counts across all experimental configurations. The performance curves demonstrate that k=3 achieves comparable UG-Mean values to k=50 in most scenarios, with differences typically within 2-3 percentage points. This pattern suggests that the proposed method effectively extracts relevant consensus information from minimal evidence sets, with additional evidences providing marginal improvements that may not justify increased computational costs.

The consistency of this finding across diverse datasets and model architectures indicates that the evidence sufficiency phenomenon reflects fundamental properties of the proposed method rather than dataset-specific or task-specific characteristics.The embedding space organization appears to concentrate the most informative evidence samples within the immediate neighborhood, rendering extended evidence gathering redundant for decision-making purposes. The robustness of minimal evidence effectiveness across experimental conditions supports the practical deployment of the framework with conservative k settings, reducing both computational overhead and potential noise from distant neighbors while maintaining decision quality comparable to more extensive evidence gathering approaches.

% Notably, higher k values occasionally introduce slight performance degradation, particularly evident in FC suppression curves where k=30 and k=50 configurations sometimes exhibit elevated false certain rates compared to k=3 and k=10. This suggests that excessive evidence aggregation may dilute the signal from highly relevant neighbors by incorporating samples with weaker semantic relationships to the test instance.

\section{Conclusion} \label{sec:Conclusion}

A central limitation of applying threshold on PE was identified: because a fixed threshold on PE couples coverage and reliability, lowering the threshold suppresses FC at the cost of sharply reduced true TC and inflated deferrals (FU), whereas raising the threshold increases TC but risks unsafe growth in FC. As a result, performance is brittle with respect to the threshold and varies substantially across datasets and backbones.

Accordingly, in this study an evidence-retrieval mechanism for uncertainty-aware decision-making was proposed and evaluated. For each test instance, proximal exemplars in an embedding space were retrieved, their predictive distributions were fused, and the certainty tag was issued by checking the consistency of prediction uncertainty behavior between the instance and its evidences. Across CIFAR-10/100 with BiT and ViT backbones, the mechanism achieved higher or comparable UG-Mean while materially reducing FC and maintaining sustainable FU, thereby improving the reliability–coverage trade-off relative to entropy thresholding. Correct certain decisions were preserved when evidential neighbors agreed with the prediction, while heterogeneous or disagreeing neighborhoods appropriately triggered uncertainty.

Beyond aggregate metrics, a practical benefit was observed in transparency: because decisions are grounded in retrieved exemplars, the supporting evidence can be inspected directly, which facilitates auditing and human oversight in safety-critical settings. The method thus provides both operational gains (lower FC at useful TC) and interpretability (traceable evidential support).

Furthermore, ablation studies clarified parameter behavior. The belief threshold primarily controlled the reliability–coverage trade-off: FC decreased monotonically as the threshold increased, TC decreased accordingly, and UG-Mean exhibited a single peak in the mid-range, which offered the best balance across datasets. The evidence-set size \(k\) had a milder effect; performance was stable for \(k\) in the \(3\)–\(30\) range.

While the evidence set contains examples with known ground truth, these true labels are intentionally excluded from the decision-making process. The current approach maintains methodological consistency by relying solely on predictive evidence and associated uncertainties. Future work could explore incorporating true labels through conformal prediction frameworks, where nonconformity scores derived from ground truth comparisons could provide additional calibration mechanisms for uncertainty quantification. However, the ground truth labels of the evidence set do serve a valuable purpose during the human review process, providing additional context and validation for cases flagged as uncertain.

Other future directions includes: (i) integrating calibrated uncertainty estimators and out-of-distribution detectors to strengthen behavior under distribution shift; (ii) studying alternative similarity metrics and learned retrieval embeddings; (iii) improving scalability via efficient indexing and incremental evidence stores; and (iv) user-facing studies that assess how evidential displays influence human trust, workload, and safety outcomes. Taken together, these directions are expected to further enhance the reliability, transparency, and deployability of evidence-based uncertainty-aware decision systems.

\bibliographystyle{IEEEtran}
\bibliography{Refs}

\vfill

\end{document}